\documentclass[10pt,twocolumn]{article}
\usepackage{marvosym}
\usepackage{newtxtext,newtxmath}
\usepackage[a4paper,left=1.5cm,right=1.5cm,top=2.2cm,bottom=2.2cm,columnsep=0.7cm]{geometry}

\usepackage[T1]{fontenc}
\usepackage[utf8]{inputenc}
\usepackage{amsmath}

\usepackage{amssymb}
\usepackage{graphicx}
\usepackage[table]{xcolor}

\usepackage[numbers,sort&compress,super]{natbib}
\usepackage{microtype}
\usepackage{caption}
\usepackage{xurl}
\usepackage{seqsplit}
\usepackage{orcidlink}
\usepackage{booktabs}
\usepackage{longtable}
\usepackage{array}
\usepackage{fvextra}
\usepackage[most]{tcolorbox}
\usepackage{enumitem}
\usepackage{titlesec}
\usepackage{fancyhdr}

\usepackage[normalem]{ulem}

\hypersetup{colorlinks=true,linkcolor=blue!70!black,citecolor=blue!70!black,
            urlcolor=blue!70!black,breaklinks=true}

\setlist{nosep}
\DeclareCaptionLabelFormat{nmi}{\textbf{Fig. #2}}
\DeclareCaptionLabelSeparator{nmi}{\textbf{ $|$ }}
\titleformat{\section}{\large\bfseries}{\thesection}{0.6em}{}
\titleformat{\subsection}{\normalsize\bfseries}{\thesubsection}{0.6em}{}
\titleformat{\subsubsection}{\normalsize\itshape}{\thesubsubsection}{0.6em}{}
\titlespacing{\section}{0pt}{12pt plus 2pt minus 2pt}{4pt plus 1pt}
\titlespacing{\subsection}{0pt}{8pt plus 2pt minus 2pt}{3pt plus 1pt}
\graphicspath{{../figures/}}
\fancypagestyle{firstpage}{%
  \fancyhf{}%
  \fancyfoot[R]{\footnotesize\thepage}%
}
\newtcolorbox{promptbox}[1]{enhanced,breakable,
  colback=black!2,colframe=black!25,colbacktitle=black!6,coltitle=black,
  boxrule=0.4pt,arc=1.5mm,left=3mm,right=3mm,top=2mm,bottom=2mm,
  fonttitle=\small\bfseries,title={#1},before skip=8pt,after skip=8pt}
\DefineVerbatimEnvironment{PromptText}{Verbatim}{fontsize=\footnotesize,
  breaklines=true,breakanywhere=true,breaksymbolleft={},breakindent=0pt}

\begin{document}
\thispagestyle{firstpage}
\twocolumn[
\vspace*{-0.5cm}
{\noindent\small\textbf{Preprint}\par}
\vspace{12pt}

{\noindent\fontsize{22}{26}\selectfont\bfseries
Potential emergence in LLM agents:\\
a statistical physics perspective\par}
\vspace{14pt}

{\noindent\large
Zhuo-Yang Song~\orcidlink{0009-0001-9727-7908}\textsuperscript{1}
\quad Hua Xing Zhu\textsuperscript{1,2}\par}
\vspace{6pt}

{\noindent\footnotesize
\textsuperscript{1}School of Physics, Peking University, Beijing 100871, China\\[2pt]
\textsuperscript{2}Center for High Energy Physics, Peking University, Beijing 100871, China\\[4pt]
e-mail: \href{mailto:zhuoyangsong@stu.pku.edu.cn}{zhuoyangsong@stu.pku.edu.cn};
\href{mailto:zhuhx@pku.edu.cn}{zhuhx@pku.edu.cn}\par}

\vspace{10pt}
\noindent\rule{\textwidth}{0.4pt}
\vspace{8pt}

{\small\noindent
There has been significant recent interest in understanding the behaviour of LLM agents from the perspective of human preferences or corpus statistics. In this paper we propose agent states as intermediate variables through which post-training can be measured and predicted.
  From a statistical physics perspective, we study the emergence of a global potential which governs the transition of state in LLM agent reasoning. We introduce a partition function to model the ideal KL-regularised transition distribution and track potential structure, mean reward and state entropy. We demonstrate our approach in two Qwen3.5-9B tasks: safety-test selection and physics-reasoning auditing. We observe that the potential structure strengthens during post-training as predicted from our framework, while joint changes in reward and state entropy provide consistency evidence. States thus provide a quantitative description of how post-training organises agent behaviour.\par}
\vspace{8pt}
\noindent\rule{\textwidth}{0.4pt}
\vspace{16pt}
]

\section{Introduction}

Large language models (LLMs) \cite{vaswani2017attention,brown2020language,kaplan2020scaling,hoffmann2022training} are becoming an increasingly important new tool for scientific discovery, experimental design and software development \cite{ref23,ref24,bran2023chemcrow,funsearch,alphaevolve,lu2024aiscientist,Song:2025pwy,Cao:2025shc,song2025iteratedagentsymbolicregression}. For such tasks, they are placed in agent frameworks with external environments, tools and memory \cite{xi2023rise,wang2024survey,Mohammadi_2025,sapkota2025aiagentsvsagentic}, and execute through reasoning-action cycles \cite{yao2023reactsynergizingreasoningacting,schick2023toolformerlanguagemodelsteach,shinn2023reflexion,wang2023voyageropenendedembodiedagent,Park2023}. Their outputs are therefore both token sequences and transitions between agent states, the task stages an agent moves through, such as the next test to run or the next reasoning checkpoint \cite{Markov2006ExtensionOT}, linking language generation to agent execution on tasks. Related work spans multi-step reasoning \cite{NEURIPS2022_9d560961,kojima2022large,wang2023selfconsistencyimproveschainthought,wang2023planandsolvepromptingimprovingzeroshot,lightman2024verify}, program-aided problem solving \cite{chen2023programthoughtspromptingdisentangling,pmlr-v202-gao23f}, reasoning-path search \cite{yao2023treethoughtsdeliberateproblem,Besta_Blach_Kubicek_Gerstenberger_Podstawski_Gianinazzi_Gajda_Lehmann_Niewiadomski_Nyczyk_Hoefler_2024}, embodied control \cite{liang2023codepolicieslanguagemodel} and open-ended continual exploration \cite{wang2023voyageropenendedembodiedagent}. What models choose, how they search and why they stagnate are increasingly difficult to understand from human intuition alone. Independent runs of the same agent on the same task may settle into different “idea basins”, with performance gaps persisting even after substantial additional computation \cite{ref30}; multi-agent networks also show diverse strategic behaviours \cite{ref21,ref22,kim2025sciencescalingagentsystems}. Explaining and predicting such behaviour calls for a deeper understanding of the patterns shaped by the models' own training and inference algorithms, and of the relationship between token-level optimisation objectives and state-level organisation.

Statistical physics methods have been used to study learning \cite{seung1992statistical,carleo2019machine} and language-model training dynamics \cite{liu2025neuralthermodynamiclawslarge,du2026reliability}. At the state level, previous studies have established ways to measure behaviour. Farquhar and colleagues clustered responses into semantic equivalence classes and used semantic entropy to measure uncertainty \cite{ref1,ref2}; Kim and colleagues modelled chain-of-thought (CoT) generation as a metastable Markov process in which each state represents a logical assertion and transitions correspond to reasoning steps \cite{ref20}. Song and colleagues modelled LLM inputs and outputs as Markov transitions and characterised a potential structure in each agent's state space \cite{ref7}. These studies quantify semantic uncertainty and state-transition structure, but the mechanisms by which training shapes them remain unclear.

Post-training adapts pretrained models to preferences and tasks through reinforcement learning from human or AI feedback, direct preference optimisation and verifiable rewards \cite{christiano2017deep,ziegler2019fine,stiennon2020learning,ouyang2022training,bai2022training,bai2022constitutional,rafailov2023direct,shao2024deepseekmath,guo2025deepseek}. Post-training can change output diversity, preference concentration and the probability of existing reasoning patterns \cite{ref3,ref4,ref26}; semantic entropy can also collapse during on-policy interaction \cite{ref5,ref6}. Hu and colleagues used evolving concept networks to explain slow thinking and policy collapse from the organisation of reasoning paths within individual responses \cite{hu2025conceptnetwork}. Yet a quantitative link between post-training objectives and state-level behaviour is missing.

\begin{figure*}[!t]
\centering
\includegraphics[width=\textwidth]{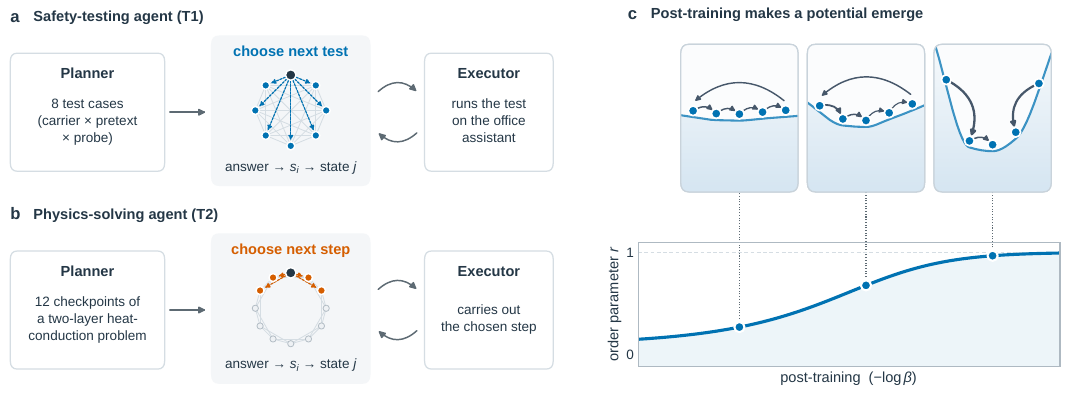}
\caption{Agent task design and illustration of potential emergence during post-training. \textbf{a}, A safety-testing agent designs a test set, T1 selects the next test, and the selected test is then executed; T1 contains 8 states, with transitions allowed between any two distinct states. \textbf{b}, A physics problem-solving agent designs a sequence of solution steps, T2 selects the next step, and the corresponding operation is then executed; T2 contains 12 states, each connected to its two neighbours in either direction on a ring. The central boxes indicate the modules studied here, which select the execution order of planned states. \textbf{c}, Schematic of potential emergence under \(T_{ij}=Q_{ij}e^{R_{ij}/\beta}/Z_i(\beta)\) for five states. Bottom, order parameter \(r\) versus post-training progress \(-\log\beta\) (\(\beta\) is the KL temperature). Top, net flow (arrows) at the three marked points: as \(\beta\) decreases, a potential (node height) increasingly sets the direction of the net flow, which changes from a loop flow on a flat landscape to a flow down the potential.}
\label{fig:task-design-theory}
\end{figure*}

We address this gap by using agent states as intermediate variables for training, with a partition-function formulation on a finite state space that permits quantitative analysis \cite{ref7}, and study how potential landscapes emerge in two tasks: sequential selection of safety tests (T1) and physics reasoning audits (T2). The main discovery is that a potential estimated without reward information explains the transition probabilities progressively better during post-training, with the observed state dependence matching theoretical predictions; joint changes in mean reward and state entropy provide consistency evidence. To test generality across training and inference conditions, we vary the KL coefficient, inference mode, sampling temperature, prompt variants, model size and family, and reward structure. Our work connects token-level optimisation to the organisation of choices and transitions in agent state spaces. In this paper we focus on post-training with fixed state representations and simplified tasks (Fig.~\ref{fig:task-design-theory}); comparisons among different pre-training and post-training settings, and the formation of state spaces, remain to be explored in future work.

\section{Results}

\subsection{Partition function on the state space}

When there are well-defined states in generic reinforcement learning problems \cite{sutton2018reinforcement}, a given reference policy and reward function allow the optimal solution to the KL-regularised objective to be expressed as an exponential reweighting of the reference transition kernel \cite{ref9,ref10,ref11,ref19}. LLMs instead generate unstructured token sequences whose state space is not obviously defined; Song and colleagues map token sequences with a parser to finite semantic states, so each valid generation corresponds to a state transition, and highly diverse responses are aggregated into a small number of states whose transition counts can be directly estimated \cite{ref7}. We use this formulation with assumptions in Section~\ref{sec:theory}. Given source state \(i\) and input \(x_i\), the parser \(s_i\) maps a valid response \(y\) to destination \(j\in\mathcal L(i)\), where \(\mathcal L(i)\) is the set of valid destinations from \(i\). In the theoretical treatment, we assume that every response maps to a valid destination. Invalid responses are handled separately in Section~\ref{sec:evaluation}. With response distributions \(p_{\boldsymbol\theta}(y\mid x_i)\) and \(p_{\mathrm{ref}}(y\mid x_i)\), the corresponding transition kernels are

\begin{equation}
T_{ij}=\sum_{y:s_i(y)=j}p_{\boldsymbol\theta}(y\mid x_i),\quad
Q_{ij}=\sum_{y:s_i(y)=j}p_{\mathrm{ref}}(y\mid x_i).
\label{eq:transition-kernel}
\end{equation}

Write \(T_{i\circ}=(T_{ij})_{j\in\mathcal L(i)}\) and \(Q_{i\circ}=(Q_{ij})_{j\in\mathcal L(i)}\) for the corresponding row distributions.

During measurement, we repeatedly reset the agent to source \(i\), sample and parse responses into destinations \(j\), and normalise their counts by the total number of samples from that source to estimate \(T_{ij}\). The reference kernel \(Q\) uses the same protocol (see Section~\ref{sec:evaluation}).

Exponential reweighting under a KL penalty is standard \cite{ref9,ref10,ref11,ref19}; our contribution is to apply it on states rather than responses, so that both $T$ and $Q$ can be measured directly. KL-regularised reinforcement learning raises expected reward $\sum_{j\in\mathcal L(i)} T_{ij} R_{ij}$ while penalising deviation from a reference policy, where $R_{ij}$ is the transition reward. If the response distribution can be optimised freely, the state-space objective is to maximise \(\sum_{j\in\mathcal L(i)}T_{ij}R_{ij}-\beta D_{\mathrm{KL}}(T_{i\circ}\Vert Q_{i\circ})\), where \(\beta>0\) sets the relative reward-regularisation scale and \(D_{\mathrm{KL}}\) denotes the Kullback--Leibler divergence. The optimum is given by:
\begin{equation}\label{eq:optimal-kernel}
\begin{aligned}
T_{ij}(\beta)&=\frac{Q_{ij}\exp(R_{ij}/\beta)}{Z_i(\beta)},\\
Z_i(\beta)&=\sum_{k\in\mathcal L(i)}Q_{ik}\exp(R_{ik}/\beta).
\end{aligned}
\end{equation}
The local partition function \(Z_i\) uses \(Q\) as its measure and inserts reward through exponential weights, as in KL-regularised control, maximum-entropy reinforcement learning and probabilistic inference \cite{ref9,ref10,ref11,haarnoja2018soft,ref19}. Varying the KL strength links reinforcement-learning stages through ideal optima, analogous to quasistatic annealing. Here, \(\beta\) acts as a temperature, with negative reward as energy \cite{jaynes1957information}. As \(\beta\to\infty\), the kernel approaches \(Q\); as \(\beta\to0\), probability concentrates on the highest-reward destinations within the support of \(Q\). We therefore use \(\beta\) to parameterise the post-training trajectory below, plotting \(-\log\beta\) so that post-training progresses from left to right. The derivation and assumptions are given in Section~\ref{sec:theory}.

We test this description in two simplified tasks: T1 selects the next safety test and T2 the next physics-reasoning checkpoint. They contain 8 and 12 predefined states, respectively (Fig.~\ref{fig:task-design-theory}a,b; definitions in Appendix~\ref{app:prompts}). The state spaces are deliberately kept simple and free from prespecified corpus bias, allowing post-training signals to be sampled explicitly. Both tasks start from Qwen3.5-9B \cite{qwen35_2026} and train for 8 rounds with LoRA \cite{hu2022lora} and PPO, using REINFORCE with a mean-reward baseline to estimate advantages \cite{schulman2017ppo,williams1992reinforce} (Table~\ref{tab:post-training-config}). Let \(\beta_0\) be the fixed relative penalty coefficient. The reference is refreshed between rounds; if every round reaches the ideal optimum, the successive reweightings combine into one relative to the initial kernel \(Q\), with \(\beta_{\mathrm{nominal},m}=\beta_0/m\). The main experiments use \(\beta_0=0.375\).

Actual training involves only a finite number of updates, so we describe the measured distributions using the effective scale in equation~\eqref{eq:beta-calibration}:
\begin{equation}\label{eq:beta-calibration}
\beta_{\mathrm{eff},m}^{(t)}=c_t\beta_{\mathrm{nominal},m},\qquad
 t\in\{\mathrm{T1},\mathrm{T2}\}.
\end{equation}
We calibrate this scale from the reward difference and Jeffreys divergence between two rounds for each source state, fitting the three training seeds of each task jointly, which gives \(c_{\mathrm{T1}}=3.04\) and \(c_{\mathrm{T2}}=3.45\) (Fig.~\ref{fig:reward-entropy}a). Main-experiment comparisons of potential, reward and entropy use this scale; the calibration and uncertainty estimates are described in Sections~\ref{sec:theory} and~\ref{sec:evaluation}.

\subsection{Measuring potential emergence in real tasks}

The emergence of capabilities in language models is widely discussed \cite{ref27,schaeffer2023emergent}. Using a reference kernel $Q$ before task-specific post-training, we show that post-training can encode existing rules and strategies in a global potential by reorganising state transitions \cite{ref7}. Relative to the reference kernel of the initial model, we track how this structure evolves along the post-training trajectory as $\beta$ varies and test whether partition function theory explains the mechanism behind the potential structure measured by Song and colleagues \cite{ref7}.

Common rewards depend on where a transition ends rather than on the path taken, as when a multi-hop and a single-hop reasoning path that reach the same correct answer receive the same reward; to analyse the mechanism of potential emergence, we therefore decompose the reward so that this destination term appears explicitly:
\begin{equation}\label{eq:reward-decomposition}
R_{ij}=h_i+u_j+\epsilon_{ij}.
\end{equation}
Here, \(h_i\) is a source-state reward offset that cancels under row normalisation and does not affect the optimal transition kernel, \(u_j\) is a destination-state reward shared across sources, and \(\epsilon_{ij}\) is a residual term. For rewards with \(\epsilon_{ij}=0\), which we call shared destination rewards, the theoretical potential is 

\begin{equation*}
\begin{aligned}
\Phi_i(\beta)&:=-u_i-\beta\log\left(\sum_{k\in\mathcal L(i)}Q_{ik}e^{u_k/\beta}\right),\\
&\xrightarrow{\beta\to0}-u_i-\max_{\substack{k\in\mathcal L(i)\\Q_{ik}>0}}u_k\\
&=:\Phi_i(0).
\end{aligned}
\end{equation*}
This potential is jointly determined by the state's reward and the highest reward among the reachable states, with the latter entering through the partition function.

For each pair of distinct states that permits transitions in both directions, we choose one fixed direction and count the pair once, denoting the resulting edge set by \(\mathcal E_*\), and let \(y_{ij}=\log(T_{ij}/T_{ji})\). For a fixed reference kernel and \(\mathcal E_*\),
\begin{equation}\label{eq:potential-limit}
\beta y_{ij}\longrightarrow\Phi_i(0)-\Phi_j(0),\qquad\beta\to0.
\end{equation}
The derivation is given in Section~\ref{sec:theory}.
Thus, in the limit \(\beta\to0\), the scaled log ratios of forward and reverse transition probabilities are determined by differences in the theoretical potential \(\Phi(0)\). The emergence of this potential function shows that post-training not only changes model behaviour continuously, but also introduces a global potential structure on the state space, providing a post-training mechanism for the tendency towards detailed balance \cite{ref7,schnakenberg1976network}. Detailed derivations of the reward decomposition, the log ratios and their limits are given in Section~\ref{sec:theory}.

Following the minimum-action method \cite{ref7}, we estimate the potential by minimising the action for each round's transition kernel, obtaining a centred potential \(\widehat\phi\) without using reward information. We then define the order parameter by the Pearson correlation in equation~\eqref{eq:order-parameter}:
\begin{equation}\label{eq:order-parameter}
r=\operatorname{corr}_{(i,j)}\!\left(\widehat\phi_i-\widehat\phi_j,
\log\frac{\widehat T_{ij}}{\widehat T_{ji}}\right)
\end{equation}
The order parameter quantifies the agreement between potential differences and the log ratios \(y_{ij}\). The correlation uses only edges in \(\mathcal E_*\) observed in both directions. At small \(\beta\),
\begin{equation}\label{eq:r-asymptotic}
r\simeq(1+\kappa+\theta\beta^2)^{-1/2}.
\end{equation}
Here \(\kappa\) is the contribution of the antisymmetric reward residual that cannot be written as a potential difference; for shared destination rewards, \(\kappa=0\); equation~\eqref{eq:r-asymptotic} then gives \(r\to1\) as \(\beta\to0\). The effective parameter \(\theta\) captures the non-potential component of the reference kernel. The derivation and validity conditions are given in Section~\ref{sec:theory}.

\begin{figure*}[!t]
\centering
\includegraphics[width=\textwidth]{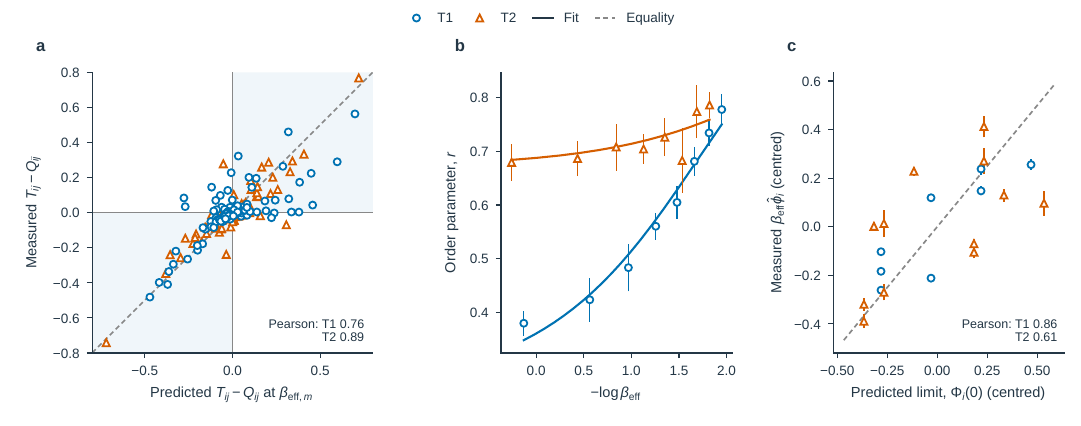}
\caption{Strengthening of potential structure during post-training and tests of its theoretical description. \textbf{a}, Change of each transition probability from the initial kernel, \(T_{ij}-Q_{ij}\): prediction of equation~\eqref{eq:optimal-kernel} at the calibrated \(\beta_{\mathrm{eff},m}\) versus measurement, for all transitions of T1 and T2. Rounds 1, 4 and 8 of the main experiment are shown; the Pearson correlations use all 8 rounds. Shading marks agreement in sign, and the dashed line denotes equality. \textbf{b}, Order parameter \(r\) for T1 and T2 as a function of calibrated \(\beta_{\mathrm{eff}}\). Solid lines show fits of equation~\eqref{eq:r-asymptotic} with \(\kappa=0\). Open markers denote measurements; error bars in b and c combine the 95\% interval from 500 bootstrap resamples with the spread over three training seeds (Section~\ref{sec:evaluation}). \textbf{c}, Centred measured potential \(\beta_{\mathrm{eff},8}\widehat\phi_i\) in round 8 versus the centred theoretical limit \(\Phi_i(0)=-u_i-\max_{k:Q_{ik}>0}u_k\). Each point represents one state. The dashed line denotes the theoretical equality \(\beta_{\mathrm{eff},8}\widehat\phi_i=\Phi_i(0)\). The Pearson correlations between the centred theoretical limit and centred measured potential are 0.86 for T1 and 0.61 for T2. The value of \(\beta_{\mathrm{eff},8}\) is taken from the calibration in Fig.~\ref{fig:reward-entropy}. We draw 4,096 samples per source per round. Panel b includes all 8 rounds.}
\label{fig:potential-emergence}
\end{figure*}

From the initial kernel alone, equation~\eqref{eq:optimal-kernel} predicts each change \(T_{ij}-Q_{ij}\) and agrees with the measured change, with Pearson correlations of 0.76 for T1 and 0.89 for T2 (Fig.~\ref{fig:potential-emergence}a).

Figure~\ref{fig:potential-emergence}b shows that post-training strengthens agreement between the log ratios and global potential differences, supporting the predicted potential structure. For shared destination rewards, we fix \(\kappa=0\) and jointly fit \(\theta_t\) and \(\beta_{i,t}\) using \(\beta_{\mathrm{fit},m,t}^{-1}=\beta_{\mathrm{eff},m,t}^{-1}+\beta_{i,t}^{-1}\) and theoretical-remainder weights (Section~\ref{sec:evaluation}); \(\beta_{i,t}\) absorbs the nonzero baseline \(r\) of the reference kernel, potentially from Qwen’s own post-training.

The measured potential in round 8, \(\beta_{\mathrm{eff},8}\widehat\phi_i\), varies across states in the same direction as the theoretical limit \(\Phi_i(0)\), with Pearson correlations of 0.86 for T1 and 0.61 for T2 (Fig.~\ref{fig:potential-emergence}c). Using the calibration in Fig.~\ref{fig:reward-entropy}, this agreement supports the predicted state dependence.

\subsection{Consistency tests of reward and state entropy}

Changes in model behaviour can appear as new organisation or as relations between continuous performance metrics. Mean reward, \(R_i=\sum_{j\in\mathcal L(i)}T_{ij}R_{ij}\), measures alignment with the training objective, whereas state Shannon entropy, \(H_{\mathrm{parser},i}=-\sum_{j\in\mathcal L(i)}T_{ij}\log T_{ij}\), measures choice dispersion \cite{cover2006information}; their joint variation could test the consistency of the partition function description. At small \(\beta\), when transitions concentrate on highest-reward destinations, expansion of the partition function gives the approximate relations
\begin{equation}
\begin{aligned}
R_i(\beta)&\simeq R_{\mathrm{max},i}-A_i e^{-\Delta_i/\beta},\\
H_{\mathrm{parser},i}(\beta)&\simeq H_{*,i}+\frac{A_i}{\beta}e^{-\Delta_i/\beta}.
\end{aligned}
\label{eq:reward-entropy-low-beta}
\end{equation}
Here, \(R_{\mathrm{max},i}\) is the maximum reward, \(\Delta_i\) is the gap between the highest and second-highest rewards, and \(H_{*,i}\) is the entropy of the reference kernel after normalisation within the set of destinations attaining the maximum reward. The same amplitude \(A_i\) controls the leading correction to both curves. Partition function identities, expansions by reward value and corrections for non-uniform reference kernels are given in Section~\ref{sec:theory}.

\begin{figure*}[!t]
\centering
\includegraphics[width=\textwidth]{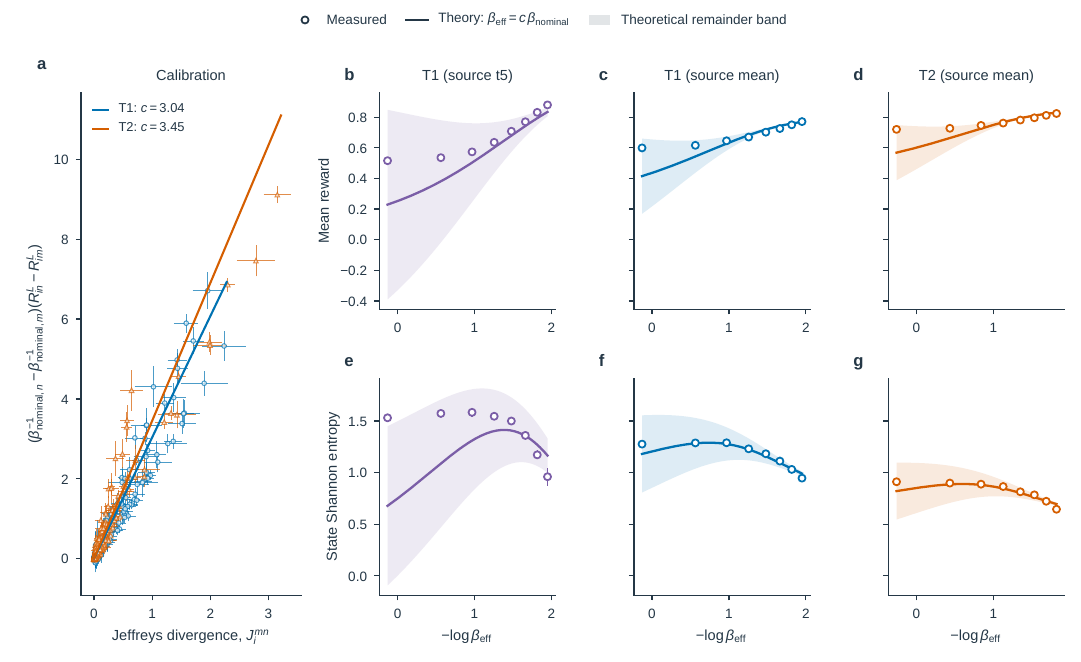}
\caption{Consistency tests of mean reward and state Shannon entropy during post-training. \textbf{a}, Proportional calibration of the effective scale. The horizontal axis shows the Jeffreys divergence between distributions from two rounds for the same source state; the vertical axis shows the corresponding reward difference multiplied by the difference in nominal inverse \(\beta\). Solid lines are fits through the origin. All 28 round pairs for each of the 8 T1 and 12 T2 source states are included, with circles and triangles distinguishing the tasks. \textbf{b,e}, Representative T1 source state \(t5\). \textbf{c,f}, Averages over the 8 T1 source states. \textbf{d,g}, Averages over the 12 T2 source states. The top row shows mean reward and the bottom row shows state Shannon entropy. Open markers denote measurements; solid lines evaluate the theoretical expressions at the calibrated \(\beta_{\mathrm{eff},m}=c_t\beta_{\mathrm{nominal},m}\). In b,e, \(R_{\mathrm{max},i}\), \(\Delta_i\) and \(H_{*,i}\) are determined by the predefined task, and only \(A_i\) is fitted with \(c_t\) fixed. In c,f and d,g, the parameters \(R_{\mathrm{max}},H_*,A,\Delta\) are jointly fitted for each task with its calibrated \(c_t\) fixed. Shading denotes theoretical remainder bands; their construction is described in Section~\ref{sec:evaluation}. Error bars in all panels combine the 95\% paired count bootstrap interval with the spread over three training seeds (Section~\ref{sec:evaluation}). We draw 4,096 samples per source per round.}
\label{fig:reward-entropy}
\end{figure*}

Figure~\ref{fig:reward-entropy}b,e present a joint test for the representative T1 state \(t5\): we fix the task-determined \(R_{\mathrm{max},i},\Delta_i,H_{*,i}\) and the calibrated \(c_t\), and fit only \(A_i\). Figure~\ref{fig:reward-entropy}c,f and d,g show averages over source states for T1 and T2, respectively. As reward gaps differ across sources, we approximate the effective task-level values using a single effective gap. The fitting parameters \(R_{\mathrm{max}},H_{*},A,\Delta\) denote, respectively, the effective reward and entropy baselines of the fitted curves as \(\beta\to0\), the shared correction amplitude and the effective reward gap.

Both tasks exhibit increasing reward, with state entropy decreasing at later stages. For the representative state, measurements agree with the leading asymptotic curves at small \(\beta\), whereas deviations at larger \(\beta\) reflect the range of validity of the small-\(\beta\) expansion. All rounds are included in the fit and weighted according to the theoretical remainder, giving the small-\(\beta\) region greater influence. The procedure is detailed in Section~\ref{sec:evaluation}.

\subsection{Inference sensitivity and model size and family comparisons}

\begin{figure*}[!t]
\centering
\includegraphics[width=\textwidth]{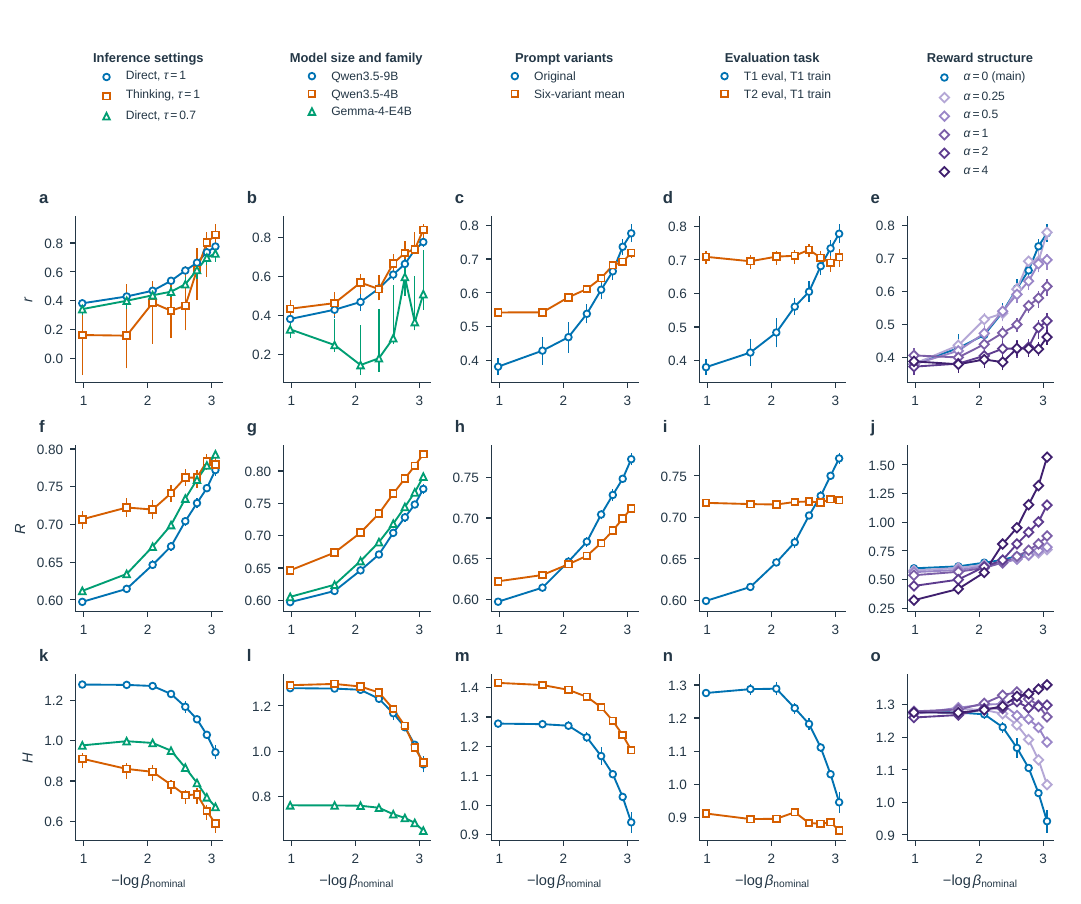}
\caption{Sensitivity of post-training organisation and reward ablation. The five columns compare inference mode and sampling temperature, model size and family, prompt variants, evaluation of the same T1-trained checkpoints on T1 and T2, and reward structure, \(R_{ij}=u_j+\alpha\epsilon_{ij}\) with shared destination rewards (\(\alpha=0\), main experiment) and an added edge-specific residual (\(\alpha>0\), diamonds). \textbf{a--e}, Order parameter \(r\); \textbf{f--j}, mean reward \(R\); \textbf{k--o}, state Shannon entropy \(H\). All 8 rounds are plotted against \(-\log\beta_{\mathrm{nominal},m}\), where \(\beta_{\mathrm{nominal},m}=0.375/m\). Open markers denote measurements; error bars show 95\% interval of 500 bootstrap resamples. Some error bars are smaller than the markers. For the main T1 protocol (blue circles), error bars also include the spread over three training seeds (Section~\ref{sec:evaluation}). For prompt variants, statistics are computed per resample and averaged equally across variants before forming the intervals. Direct-answer and thinking-mode evaluations use 4,096 and 256 samples per source, respectively.}
\label{fig:ablations}
\end{figure*}

The results above centre on potential-structure emergence, with reward and entropy changes providing consistency evidence. To test its dependence on conditions, we vary inference mode, sampling temperature, model size and family, prompt variants and reward structure in T1, and also evaluate the same T1-trained checkpoints on T1 and T2. Figure~\ref{fig:ablations} shows these comparisons in columns and the order parameter \(r\), mean reward \(R\) and state Shannon entropy \(H\) in rows. All use \(\beta_{\mathrm{nominal},m}=0.375/m\) to align progress.

First, we hold the Qwen3.5-9B T1 adapters fixed and switch evaluation from direct-answer to thinking mode (Fig.~\ref{fig:ablations}a,f,k). Both retain the post-training trends of increasing order parameter \(r\) and reward and decreasing entropy. Thinking mode already gives the initial model higher reward and lower entropy, showing that inference changes its state distribution. Thinking-mode evaluation uses 256 samples per source.

For the same checkpoints, we reduce the direct-answer sampling temperature \(\tau\) from 1 to 0.7 (Fig.~\ref{fig:ablations}a,f,k). The overall trends of increasing reward, decreasing entropy at later stages and increasing \(r\) persist at the lower temperature. Lowering the temperature concentrates choices and raises mean reward but does not produce a higher \(r\), showing that choice concentration and the agreement of the log ratios \(y_{ij}\) with a potential are distinct observable features.

Second, we train Qwen3.5-4B \cite{qwen35_2026} for 8 rounds from its initial checkpoint using the same T1 post-training protocol (Fig.~\ref{fig:ablations}b,g,l). Increases in \(r\) and reward are observed at both model sizes, and the trajectories of decreasing state entropy are similar. Training Gemma-4-E4B with the same protocol also increases reward, from 0.61 to 0.79, and decreases state entropy; its order parameter rises from 0.10 in the initial model to 0.51 in round 8, with wide intervals. The joint variation of the three observables indicates that the theory is not restricted to a single model size or model family.

We then evaluate the same 9B adapters with six prompt variants not used for training or model selection (Fig.~\ref{fig:ablations}c,h,m). They comprise rewording, answer-code permutation, both changes and three related application scenarios, while retaining the abstract state graph, destination rewards and parsing rules. Across variants, mean order parameter \(r\) and reward increase and mean state entropy decreases, although the changes are smaller than for the original prompts. The resulting organisation therefore generalises to unseen task formulations.

We also evaluate the same T1-trained checkpoints on T1 and T2 (Fig.~\ref{fig:ablations}d,i,n). On T1, \(r\) and reward increase and state entropy decreases as in the main experiment, whereas the corresponding T2 changes are negligible. This cross-task evaluation indicates that the observed post-training changes are specific to the trained space.

Finally, we retain the T1 state graph and training procedure but add edge-specific rewards, \(R_{ij}=u_j+\alpha\epsilon_{ij}\) with \(\alpha=0.25\)--4 (Fig.~\ref{fig:ablations}e,j,o). Every row and column of \(\epsilon_{ij}\) sums to zero; it is defined in equation~\eqref{eq:app-edge-epsilon} in Appendix~\ref{app:prompt-t1}. Post-training still increases mean reward at every \(\alpha\), but as the edge-specific term outweighs the shared destination term \(u_j\), the order parameter in round 8 decreases from 0.78 to 0.46 and, for \(\alpha\geq1\), state entropy no longer decreases. This comparison shows that, under the present task and training protocol, reward improvement alone is insufficient to produce the same organisation by a potential and state concentration. Whether the reward contains a destination term shared across sources affects the outcome.

These comparisons test sensitivity to inference mode, model size and family, and task formulation; the reward-structure control supports the role of the shared destination term in potential-structure emergence.

\subsection{Training sensitivity}

\begin{figure*}[!t]
\centering
\includegraphics[width=\textwidth]{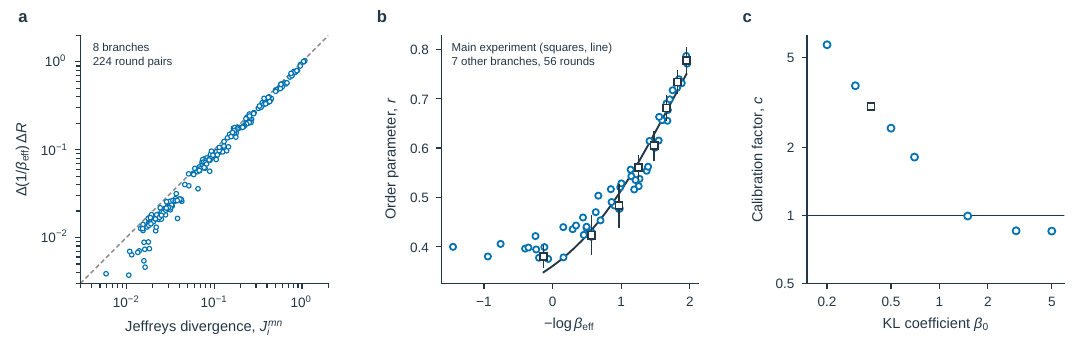}
\caption{Sensitivity to the training procedure in T1. \textbf{a}, Calibration relation of Fig.~\ref{fig:reward-entropy}a in effective units: \(\Delta(1/\beta_{\mathrm{eff}})\,\Delta R\) versus the Jeffreys divergence between two rounds, averaged over the 8 source states, for every round pair of the 8 branches. Each branch uses its own calibration factor \(c\); the dashed line denotes equality. \textbf{b}, Order parameter \(r\) of all other branches at their own calibrated \(\beta_{\mathrm{eff},m}\) (circles), with the main-experiment measurements and fit of Fig.~\ref{fig:potential-emergence}b (squares, line); no parameter is refitted. \textbf{c}, Calibration factor \(c\) versus the KL coefficient \(\beta_0\); the main experiment is the square, and the horizontal line denotes \(c=1\), the ideal optimum in every round. In b, error bars of the main experiment combine the 95\% interval from 500 bootstrap resamples with the spread over three training seeds; in c, the error bar shows the spread of \(c\) over the three seeds (Section~\ref{sec:evaluation}). Branches comprise the main experiment and seven further KL coefficients (Section~\ref{sec:experiments}); each branch is calibrated from all of its measured rounds. We draw 4,096 samples per source per round.}
\label{fig:training-ablations}
\end{figure*}

We further test training sensitivity by varying the KL coefficient during training to control \(\beta_0\). For each training branch, the effective scale is \(\beta_{\mathrm{eff},m}=c\beta_{\mathrm{nominal},m}=c\beta_0/m\), with \(c\) calibrated from all measured rounds of that branch as in the main experiment. Rescaled by its own \(c\), every branch follows the same calibration relation (Fig.~\ref{fig:training-ablations}a), and at their own \(\beta_{\mathrm{eff}}\), the order parameters of all 7 other branches lie close to the main-experiment curve of Fig.~\ref{fig:potential-emergence}b without refitting (Fig.~\ref{fig:training-ablations}b). If every round reached the ideal optimum, \(1/\beta_{\mathrm{eff}}\) would increase by \(1/\beta_0\) per round, that is, \(c=1\). With the default learning rate and steps per round, \(c\) is close to 1 for \(\beta_0\geq1.5\) (Fig.~\ref{fig:training-ablations}c). We attribute the deviation of \(c\) from 1 at smaller \(\beta_0\) to training not reaching saturation because the numbers of optimisation steps and rollouts are finite (Section~\ref{sec:experiments}). These results indicate that, across post-training settings, the deviation of finite training from the ideal optimum is captured by the single factor \(c\).

\section{Discussion}

This work uses agent states as intermediate variables to study post-training and the resulting potential-structure emergence in LLM agents. We show that a partition function links post-training to inference and that, in two simplified tasks, post-training strengthens agreement between the log ratios \(y_{ij}\) and global potential differences. Independent potential estimates and the reward-structure control provide the central evidence (Figs~\ref{fig:potential-emergence} and~\ref{fig:ablations}e,j,o); joint reward-entropy changes provide consistency evidence (Fig.~\ref{fig:reward-entropy}). The three observables therefore test distinct aspects of the description.

The three observables connect our results to work on entropy and diversity. Kirk and colleagues found that RLHF can improve generalisation while reducing output diversity \cite{ref3}. Xiao and colleagues showed that standard KL-regularised RLHF need not reproduce human preference proportions, because the reference distribution and regularisation strength jointly determine the output distribution \cite{ref4}. Cui and colleagues reported a relation between mean token-level policy entropy and downstream accuracy \cite{ref14}, while our results provide a mechanistic test of the reward-entropy connection at the state level.

Applying exponential reweighting from KL regularisation, maximum-entropy reinforcement learning and energy-based models to language-model generation \cite{ref8,ref9,ref10,ref11,ref19} lets states, so far used mainly to understand reasoning \cite{ref1,ref2,ref20,ref7}, also describe training. This perspective shows that agent behaviour can be understood by studying the training and inference mechanisms themselves. When corpus statistics and human-preference comparisons fall short, the reward structure, the reference distribution and the inference conditions serve as mechanistic clues and could inform training. Grounded in partition function theory, our approach could make testable predictions about inference and training regularities in LLM-driven agents, enriching the toolbox for studying LLMs.

Several questions remain open. Here the states are predefined, regex-extractable outputs in a finite space. In more realistic settings, they may be abstract checkpoints, execution logs or continuous multimodal observations; defining such states and expressing their rewards as \(R_{ij}\) remain challenging. Second, this work analyses how states are organised during post-training within a fixed state space, yet it leaves open whether, even given such a state space, LLM generation can cover its non-trivial regions. Yue and colleagues' work on verifiable-reward reinforcement learning highlights the distinction between changing probabilities of existing behaviours and acquiring new ones \cite{ref26}; studies of representational geometry and its relation to behaviour suggest ways to investigate this coverage \cite{ref12,ref13}. Controlled comparisons across pre-training and post-training settings are a natural next step. Finally, training is often out of equilibrium. For example, Park and colleagues show that probability-ratio clipping can affect entropy and even reduce it under random rewards \cite{ref15}. Partition function theory provides an equilibrium reference, and our effective-scale calibration captures deviations from its ideal optimum without describing full non-equilibrium dynamics. Understanding those deviations may connect training and inference behaviour more deeply.

\section{Methods}\label{sec:methods}

\subsection{Theoretical derivation}\label{sec:theory}

\paragraph{Reinforcement-learning objective.}
The KL-regularised reinforcement learning considered here aims to increase expected task reward while limiting policy deviations from a reference policy. We retain the definitions of states, parsers and rewards introduced in the Results. Denote the current policy by \(p_{\boldsymbol\theta}(y\mid i)\equiv p_{\boldsymbol\theta}(y\mid x_i)\), where \(\boldsymbol\theta\) denotes the model parameters, and let \(p_{\mathrm{ref}}(y\mid i)\) be the frozen reference policy.

For two probability distributions \(p,q\), define the Kullback--Leibler (KL) divergence as \(D_{\mathrm{KL}}(p\Vert q)=\mathbb E_{z\sim p}[\log(p(z)/q(z))]\) \cite{ref10,ref19,cover2006information}. It measures the deviation of the current distribution from the reference distribution. Let \(a>0\) be the overall task-reward scale and \(\lambda>0\) the KL penalty coefficient. The reinforcement-learning objective for source state \(i\) is given by equation~\eqref{eq:rl-objective}:
\begin{equation}\label{eq:rl-objective}
\mathcal O_i(\boldsymbol\theta)
=a\,\mathbb E_{y\sim p_{\boldsymbol\theta}(\cdot\mid i)}[R_{i,s_i(y)}]
-\lambda D_{\mathrm{KL}}\!\left(p_{\boldsymbol\theta}(\cdot\mid i)\Vert p_{\mathrm{ref}}(\cdot\mid i)\right).
\end{equation}
For \(n\) source states, the overall objective is to maximise \(n^{-1}\sum_i\mathcal O_i(\boldsymbol\theta)\). Dividing the objective by \(a\) gives the relative penalty scale \(\beta=\lambda/a\); here, \(a=1\).

\paragraph{State-space partition function.}
Let \(p(y\mid i,j)\) denote the within-class distribution conditional on \(s_i(y)=j\). The chain rule for KL divergence applied to responses grouped by their parsed class gives \cite{cover2006information}
\begin{equation*}
\begin{aligned}
D_{\mathrm{KL}}(p_{\boldsymbol\theta}\Vert p_{\mathrm{ref}})
&=D_{\mathrm{KL}}(T_{i\circ}\Vert Q_{i\circ})\\
&\quad+\sum_{j\in\mathcal L(i)}T_{ij}D_{\mathrm{KL}}\!\left(p_{\boldsymbol\theta}(y\mid i,j)\Vert p_{\mathrm{ref}}(y\mid i,j)\right)\\
&\geq D_{\mathrm{KL}}(T_{i\circ}\Vert Q_{i\circ}).
\end{aligned}
\end{equation*}
The source-state condition \(i\) is omitted on the left. Redistribution within a class does not change task reward, so under free optimisation the within-class distributions equal the reference distributions and the second term vanishes. The objective therefore depends only on the state-transition distribution:
\begin{equation*}
\begin{aligned}
\mathcal F_i(T_{i\circ})
&=\sum_{j\in\mathcal L(i)}T_{ij}R_{ij}
-\beta\sum_{j\in\mathcal L(i)}T_{ij}\log\frac{T_{ij}}{Q_{ij}},\\
&\qquad T_{ij}\geq0,\quad\sum_{j\in\mathcal L(i)}T_{ij}=1.
\end{aligned}
\end{equation*}
Finiteness of the KL divergence requires \(T_{ij}=0\) wherever \(Q_{ij}=0\). For \(Q_{ij}>0\), introduce the normalisation multiplier \(\eta_i\). The stationarity condition
\(R_{ij}-\beta[\log(T_{ij}/Q_{ij})+1]+\eta_i=0\) gives the unique optimum
\begin{equation*}
T_{ij}^{\star}=\frac{Q_{ij}e^{R_{ij}/\beta}}{Z_i(\beta)},\qquad
Z_i(\beta)=\sum_{j\in\mathcal L(i)}Q_{ij}e^{R_{ij}/\beta}.
\end{equation*}
where \(Z_i\) ensures row normalisation; strict concavity makes this stationary point the global maximum.

In round \(m\), use \(T^{\star(m-1)}\) as the reference with fixed relative penalty \(\beta_0=\lambda/a\). If every round reaches the ideal optimum, then
\begin{equation*}
T_{ij}^{\star(m)}\propto T_{ij}^{\star(m-1)}e^{R_{ij}/\beta_0}
\propto Q_{ij}^{(0)}e^{mR_{ij}/\beta_0},\qquad Q^{(0)}=T^{\star(0)}.
\end{equation*}
We therefore define the nominal scale in round \(m\) as \(\beta_{\mathrm{nominal},m}=\beta_0/m\). The distribution in that round is thus equivalent to the result of a single optimisation against the fixed reference kernel \(Q^{(0)}\) at nominal scale \(\beta_{\mathrm{nominal},m}\). Due to the same training protocol, the relation between nominal and effective scales is unchanged across rounds, making linear calibration possible.

\paragraph{Mean reward and state entropy.}
Write \(\rho=1/\beta\) and \(\mathcal Z_i(\rho)=Z_i(1/\rho)\). This section assumes that optimisation reaches the ideal optimum. Let \(R_i=\sum_{j\in\mathcal L(i)}T_{ij}^{\star}R_{ij}\) denote mean reward and \(H_{\mathrm{parser},i}\) denote the state Shannon entropy. Substituting \(\log T_{ij}^{\star}=\log Q_{ij}+\rho R_{ij}-\log\mathcal Z_i\) gives
\begin{equation}\label{eq:entropy}
\begin{aligned}
R_i&=\partial_\rho\log\mathcal Z_i
=\frac{\sum_{j\in\mathcal L(i)}Q_{ij}R_{ij}e^{\rho R_{ij}}}{\mathcal Z_i},\\
H_{\mathrm{parser},i}&=-\sum_{j\in\mathcal L(i)}T_{ij}^{\star}\log T_{ij}^{\star}
=\log\mathcal Z_i-\rho R_i-\langle\log Q_{i\circ}\rangle_{T_{i\circ}^{\star}}.
\end{aligned}
\end{equation}
where \(\langle f_j\rangle_{T_{i\circ}^{\star}}=\sum_{j\in\mathcal L(i)}T_{ij}^{\star}f_j\).

To derive the forms of reward and entropy as \(\beta\to 0\), we retain the reward-level notation used in the Results. Let \(\mathcal M_{0,i},\mathcal M_{1,i}\) denote the destination sets attaining the highest and second-highest rewards, respectively, with maximum reward \(R_{\mathrm{max},i}\) and gap \(\Delta_i>0\). Normalising the reference kernel within each level, define the reference probability and conditional entropy of each set as
\begin{equation*}
\begin{aligned}
q_{\ell,i}&=\sum_{j\in\mathcal M_{\ell,i}}Q_{ij},\qquad \ell=0,1,\\
H_{\ell,i}&=-\sum_{j\in\mathcal M_{\ell,i}}
\frac{Q_{ij}}{q_{\ell,i}}\log\frac{Q_{ij}}{q_{\ell,i}}.
\end{aligned}
\end{equation*}
Expanding the partition function by distinct reward values gives
\begin{equation*}
Z_i(\beta)=e^{R_{\mathrm{max},i}/\beta}
\left[q_{0,i}+q_{1,i}e^{-\Delta_i/\beta}
+O(e^{-\Delta_{2,i}/\beta})\right],
\end{equation*}
where \(\Delta_{2,i}\) is the gap between the highest and third-highest rewards, set to \(\infty\) when no third distinct reward value exists. With \(b_i=q_{1,i}/q_{0,i}\) and \(\delta_i=\min(2\Delta_i,\Delta_{2,i})\),
\begin{equation*}
\log\mathcal Z_i=\rho R_{\mathrm{max},i}+\log q_{0,i}
+b_i e^{-\rho\Delta_i}+O(e^{-\rho\delta_i}).
\end{equation*}
The term \(2\Delta_i\) comes from the quadratic term in the expansion of \(\log(1+x)\), whereas \(\Delta_{2,i}\) comes from the third-highest reward value; the smaller exponent determines the order of the remainder. The reference conditional distribution is preserved within each reward level, so
\begin{equation*}
\begin{aligned}
\langle\log Q_{i\circ}\rangle
&=\log q_{0,i}-H_{0,i}\\
&\quad+b_i(\log b_i-H_{1,i}+H_{0,i})e^{-\rho\Delta_i}
+O(e^{-\rho\delta_i}).
\end{aligned}
\end{equation*}
Substitution into equation~\eqref{eq:entropy} gives
\begin{equation*}
\begin{aligned}
R_i&=R_{\mathrm{max},i}-A_i e^{-\Delta_i/\beta}+O(e^{-\delta_i/\beta}),\\
H_{\mathrm{parser},i}&=H_{*,i}
+\left(\frac{A_i}{\beta}+C_i\right)e^{-\Delta_i/\beta}
+O(\beta^{-1}e^{-\delta_i/\beta}),\\
A_i&=\Delta_i b_i,\qquad H_{*,i}=H_{0,i},\\
C_i&=b_i(1-\log b_i+H_{1,i}-H_{0,i}).
\end{aligned}
\end{equation*}
For a unique optimal destination, \(H_{*,i}=0\). The reference probability ratio \(b_i\) determines the amplitude \(A_i\), while the within-level conditional entropies also enter the next-to-leading coefficient \(C_i\). Retaining the leading terms gives equation~\eqref{eq:reward-entropy-low-beta}.

Different sources may have different reward gaps, so an equally weighted average over source states generally remains a sum of several exponential terms. To describe effective task-level values over a finite training interval, we use the single-effective-gap approximation in equation~\eqref{eq:task-mean-rh}:
\begin{equation}\label{eq:task-mean-rh}
\begin{aligned}
\bar R(\beta)&\simeq R_{\mathrm{max}}-A e^{-\Delta/\beta},\\
\bar H_{\mathrm{parser}}(\beta)&\simeq H_*
+\frac{A}{\beta}e^{-\Delta/\beta}.
\end{aligned}
\end{equation}
All four coefficients are effective fitting parameters for the task-level values.

\paragraph{Potential structure and order parameter.}
Tan and Hong provided an interpretation of potential structure in terms of an energy-based model for the special case in which the reward is specified in advance as a difference between state potentials and the reference policy is symmetric \cite{ref8}. We extend this perspective to a general reward decomposition and examine the conditions under which shared destination rewards can induce potential structure through KL-regularised post-training.

For a general reward, \(h_i,u_j\) in equation~\eqref{eq:reward-decomposition} can be determined by an equally weighted least-squares decomposition over candidate edges. Its normal equations yield the residual conditions \(\sum_{j\in\mathcal L(i)}\epsilon_{ij}=0\) and \(\sum_{i:j\in\mathcal L(i)}\epsilon_{ij}=0\). Defining \(\widetilde Z_i=e^{-h_i/\beta}Z_i\), row normalisation eliminates the source-state term:
\begin{equation*}
T_{ij}^{\star}=\frac{Q_{ij}e^{(u_j+\epsilon_{ij})/\beta}}{\widetilde Z_i},\qquad
\widetilde Z_i=\sum_{k\in\mathcal L(i)}Q_{ik}e^{(u_k+\epsilon_{ik})/\beta}.
\end{equation*}
For \((i,j)\in\mathcal E_*\), let \(a^Q_{ij}=\log(Q_{ij}/Q_{ji})\). Then
\begin{equation}\label{eq:reciprocal-decomposition}
y_{ij}=a^Q_{ij}+\frac{u_j-u_i}{\beta}
+\frac{\epsilon_{ij}-\epsilon_{ji}}{\beta}
-\log\widetilde Z_i+\log\widetilde Z_j.
\end{equation}
Define the theoretical potential at finite \(\beta\) as
\begin{equation*}
\Phi_i(\beta):=-u_i-\beta\log\widetilde Z_i(\beta).
\end{equation*}
Equation~\eqref{eq:reciprocal-decomposition} can then be written as
\begin{equation*}
\beta y_{ij}=\Phi_i(\beta)-\Phi_j(\beta)+\epsilon_{ij}-\epsilon_{ji}+\beta a^Q_{ij}.
\end{equation*}
For shared destination rewards, \(\epsilon_{ij}=0\) and \(\beta\log\widetilde Z_i\to\max_{k\in\mathcal L(i):Q_{ik}>0}u_k\), so \(\lim_{\beta\to0}\Phi_i(\beta)=\Phi_i(0)\), yielding equation~\eqref{eq:potential-limit}. Thus, the part of the log ratios that cannot be expressed as potential differences comes only from the antisymmetric reward residual and the reference kernel.

For an edge variable \(x\) on \(\mathcal E_*\), define centring by
\begin{equation*}
\widetilde x_{ij}=x_{ij}-\bar x,\qquad
\bar x=\frac{1}{|\mathcal E_*|}\sum_{(i,j)\in\mathcal E_*}x_{ij}.
\end{equation*}
The variances and covariances below are computed with equal weights over these edges. When comparing with measured \(r\), centring, variance, covariance and projection are all restricted to edges observed in both directions in that round. Let \(\widehat\phi\) be the state potential estimated by equation~\eqref{eq:potential-estimator}, and write \((\Delta\psi)_{ij}=\psi_i-\psi_j\). The order parameter is then \(r=\operatorname{corr}(\widetilde{\Delta\widehat\phi},\widetilde y)\). Following gradient projection on graphs \cite{Jiang2011graph}, define the equally weighted least-squares projection
\begin{equation*}
\psi_x\in\underset{\sum_i\psi_i=0}{\arg\min}
\sum_{(i,j)\in\mathcal E_*}
\left[\widetilde x_{ij}-\widetilde{\Delta\psi}_{ij}\right]^2,
\qquad
\Pi(\widetilde x)=\widetilde{\Delta\psi_x}.
\end{equation*}
Let \(I\) denote the identity operator and set
\begin{equation*}
g_\beta=\Pi(\beta\widetilde y),\qquad
e=(I-\Pi)\widetilde{(\epsilon_{ij}-\epsilon_{ji})},\qquad
q=(I-\Pi)\widetilde{a^Q}.
\end{equation*}
In equation~\eqref{eq:reciprocal-decomposition}, both the destination-reward and partition function terms are potential differences, so \(\beta\widetilde y=g_\beta+e+\beta q\). The first-order least-squares condition is
\begin{equation*}
\bigl[\widetilde x-\Pi(\widetilde x)\bigr]^{\mathsf T}\widetilde{\Delta\psi}=0,
\qquad\forall\psi,
\end{equation*}
Thus, the projection residual is orthogonal to every centred potential difference. Writing
\(G_\beta=\operatorname{Var}(g_\beta)\), \(E=\operatorname{Var}(e)\), \(Q_{\mathrm{circ}}=\operatorname{Var}(q)\) and \(C=\operatorname{Cov}(e,q)\), we obtain
\begin{equation*}
r_\Pi(\beta):=\operatorname{corr}(g_\beta,\beta\widetilde y)
=\left[1+\frac{E+2\beta C+\beta^2Q_{\mathrm{circ}}}{G_\beta}\right]^{-1/2}.
\end{equation*}
Furthermore, \(r=\eta_\beta r_\Pi\), where \(\eta_\beta=\operatorname{corr}(\widetilde{\Delta\widehat\phi},g_\beta)\). In the reward ablation, the reward residual is specified without reference to \(Q\); we therefore approximate its cross-covariance with the non-potential reference component as zero. With this approximation, and when \(\eta_\beta\approx1\) and \(G_\beta\approx G>0\), we obtain
\begin{equation*}
r\simeq(1+\kappa+\theta\beta^2)^{-1/2},\qquad
\kappa=E/G,\qquad \theta=Q_{\mathrm{circ}}/G.
\end{equation*}
Thus, \(\kappa\) corresponds to the part of the antisymmetric reward residual that cannot be expressed as potential differences. For a fixed reference kernel, shared destination reward and \(\mathcal E_*\), \(G_\beta=G+O(\beta)\) gives \(r_\Pi(\beta)=(1+\theta\beta^2)^{-1/2}+O(\beta^3)\).

\paragraph{Jeffreys divergence and calibration factor.}
Denote the mean reward in round \(m\) by \(R_{im}=\sum_{j\in\mathcal L(i)}T_{ij}^{(m)}R_{ij}\). Let the inverse scale \(\rho_m=\beta_{\mathrm{eff},m}^{-1}=(c_t\beta_{\mathrm{nominal},m})^{-1}\) describe the actual scale of distributional change. Then
\begin{equation*}
\begin{aligned}
T_{ij}^{(m)}&=\frac{Q_{ij}e^{\rho_mR_{ij}}}{\mathcal Z_i(\rho_m)},\qquad
\mathcal Z_i(\rho)=\sum_{j\in\mathcal L(i)}Q_{ij}e^{\rho R_{ij}},\\
\log\frac{T_{ij}^{(n)}}{T_{ij}^{(m)}}
&=(\rho_n-\rho_m)R_{ij}-\log\frac{\mathcal Z_i(\rho_n)}{\mathcal Z_i(\rho_m)}.
\end{aligned}
\end{equation*}
Multiplying the log ratio by \(T_{ij}^{(n)}-T_{ij}^{(m)}\) and summing over \(j\) cancels the partition function term through normalisation, giving the Jeffreys divergence between the distributions at the two rounds:
\begin{equation*}
\begin{aligned}
J_i^{mn}
&=D_{\mathrm{KL}}(T_{i\circ}^{(n)}\Vert T_{i\circ}^{(m)})
+D_{\mathrm{KL}}(T_{i\circ}^{(m)}\Vert T_{i\circ}^{(n)})\\
&=\sum_{j\in\mathcal L(i)}(T_{ij}^{(n)}-T_{ij}^{(m)})\log\frac{T_{ij}^{(n)}}{T_{ij}^{(m)}}\\
&=(\rho_n-\rho_m)(R_{in}-R_{im}).
\end{aligned}
\end{equation*}
This yields the difference relation used for calibration:
\begin{equation}\label{eq:calibration-proportional}
(\beta_{\mathrm{nominal},n}^{-1}-\beta_{\mathrm{nominal},m}^{-1})(R_{in}-R_{im})=c_tJ_i^{mn}.
\end{equation}

\subsection{Experimental configuration}\label{sec:experiments}

T1 and T2 measure the sequencing modules of a safety-testing agent and a physics problem-solving agent, respectively. The 8 states in T1 are defined by three binary attributes: delivery carrier, persuasion pretext and probed behaviour. Transitions are allowed between any two distinct states. T2 represents a two-layer steady-state heat-conduction problem with interfacial thermal contact resistance as 12 checkpoints, each connected to its two neighbours in either direction on a ring. The tasks have 7 and 4 valid destinations per source, respectively, and both exclude self-transitions. State definitions, candidate graphs and answer-code mappings remain fixed throughout training and evaluation.

Each input contains a task description, the current state, coded candidate destinations and response-format requirements; candidate options do not display rewards. T1 uses only a user message, whereas T2 also includes a system message specifying the role and response format. Inputs are encoded using the official Qwen3.5 chat template. Complete prompt templates, state descriptions and source-specific candidate mappings are given in Appendix~\ref{app:prompts}.

The main experiments use shared destination rewards: whenever the model transitions from source state \(i\) to a valid destination state \(j\), it receives reward \(1+R_{ij}=1+u_j\). All transitions into the same destination therefore receive the same reward, irrespective of the source. Outputs that cannot be parsed into a valid destination state receive a score of \(0\). T1 utilities encode safety-test priorities, whereas T2 utilities encode priorities for physics-auditing checkpoints; all values are predefined destination-state utilities. State meanings are given in Tables~\ref{tab:prompt-states-t1} and~\ref{tab:prompt-states-t2} in Appendix~\ref{app:prompts}, and rewards \(u_j\) are given in Tables~\ref{tab:prompt-map-t1} and~\ref{tab:prompt-map-t2}.

The two tasks are trained independently from the same \href{https://huggingface.co/Qwen/Qwen3.5-9B/tree/c202236235762e1c871ad0ccb60c8ee5ba337b9a}{fixed revision of the Qwen3.5-9B checkpoint}, retaining one 8-round trajectory per task and training seed.

Training uses the open-source implementations \href{https://github.com/OpenRLHF/OpenRLHF/tree/v0.10.4}{OpenRLHF 0.10.4} \cite{hu2025openrlhf} and \href{https://github.com/huggingface/peft/tree/v0.19.1}{PEFT 0.19.1} \cite{mangrulkar2022peft}; key settings are listed in Table~\ref{tab:post-training-config}. Each training round is launched separately with a reinitialised optimiser. From the second round onwards, the LoRA parameters saved after the preceding round are loaded to continue training. After each policy update, the updated model weights are synchronised with vLLM \cite{kwon2023vllm} for sampling, so that the next training batch is generated by the current policy. The model at the start of each round serves as the KL reference and remains frozen throughout the 4 optimisation steps in that round. The loading and synchronisation code added for Qwen3.5, together with the complete launch arguments, is available as open source with the experimental code on \href{https://github.com/SonnyNondegeneracy/partition-function-llm}{GitHub} \cite{song2026partitioncode}.

\begin{table}[htbp]
\centering
\caption{Key training settings for the T1 and T2 main experiments.}
\label{tab:post-training-config}
\small
\renewcommand{\arraystretch}{1.12}
\begin{tabular}{@{}p{0.28\linewidth}p{0.67\linewidth}@{}}
\toprule
Item & Configuration \\
\midrule
LoRA & Rank 16, scaling parameter \(\alpha_{\mathrm{LoRA}}=32\), dropout 0.05; applied to all \texttt{q\_proj} and \texttt{v\_proj} modules \\
Advantage estimation & REINFORCE with the mean reward of responses to the same prompt as a baseline; advantages are not divided by their standard deviation \\
KL coefficient & \(\lambda=0.375\), using the \(k_1\) estimator \cite{schulman2020kl}; added to the return token by token \\
Training sampling and batch size & 64 samples per source per step; total batch size 512/768 for T1/T2 \\
Updates & 4 optimisation steps per round, with 1 pass over each sampled batch; 8 rounds and 32 steps in total \\
Optimiser & AdamW \cite{kingma2015adam,loshchilov2019adamw}; constant learning rate \(10^{-5}\), no warm-up; \((\beta_1,\beta_2)=(0.9,0.95)\), \(\epsilon=10^{-8}\), weight decay 0 \\
Generation & Non-thinking mode; maximum 32 generated tokens; sampling temperature and top-p both 1 \\
Other settings & PPO clipping threshold 0.2 \cite{schulman2017ppo}, discount factor \(\gamma=1\); default PEFT initialisation for LoRA, with biases excluded from training \\
\bottomrule
\end{tabular}
\end{table}

Let \(\lambda\) denote the KL penalty coefficient in the implementation and \(a>0\) the overall multiplicative factor applied to the task reward. For a training objective of the form \(a\langle R\rangle-\lambda D_{\mathrm{KL}}\), dividing by \(a\) gives the nominal training coefficient \(\beta_0=\lambda/a\) used in the main text; here, \(a=1\).

Sensitivity and ablation experiments are based on T1. Comparisons of inference mode, sampling temperature and prompt formulation reuse checkpoints from the main experiment; the other conditions are trained independently. The factors are direct-answer versus thinking mode; evaluation temperatures \(\tau=1\) and \(0.7\) for direct answers; Qwen3.5-9B versus Qwen3.5-4B and Gemma-4-E4B; the original prompt and six variants (rewording, code permutation, both changes, customer service, coding assistance and healthcare administration); shared destination rewards versus rewards with an added edge-specific residual, \(R_{ij}=u_j+\alpha\epsilon_{ij}\) with the \(\epsilon_{ij}\) of equation~\eqref{eq:app-edge-epsilon} and \(\alpha=0.25,0.5,1,2,4\); and eight KL coefficients, \(\lambda=0.2,0.3,0.375,0.5,0.7,1.5,3,5\).

To vary the amount of training per round, we trained four further branches at \(\lambda=0.375\), with learning rate and optimisation steps per round \((2\times10^{-5},4)\), \((10^{-5},8)\) and \((5\times10^{-6},8)\) for 4 rounds and \((10^{-5},32)\) for 2 rounds; all other settings are as in the main experiment. The per-round change of \(1/\beta_{\mathrm{eff}}\) increases with the product of learning rate and steps per round and reaches \(1/\lambda\), that is, \(c\approx1\), at eight times the default product. The main protocol of each task is trained with three seeds, whose spread enters the error bars of Figs~\ref{fig:potential-emergence}--\ref{fig:training-ablations}. The KL branches, the training-amount branches and the seeds are measured with vLLM using the same prompts, parser and 4,096 samples per source. Gemma-4-E4B (\texttt{google/gemma-4-E4B-it}) is trained with the T1 protocol; because vLLM 0.22.1 resets its LoRA weights on loading, each checkpoint is measured after merging the adapter into the base weights. For a Qwen3.5-9B checkpoint, merged measurement gives \(r=0.745\), \(R=0.753\) and \(H=1.030\), compared with 0.790, 0.765 and 0.976 with the adapter, so it slightly understates the training effect.

All training uses direct-answer mode. For thinking-mode evaluation, the empty thinking block in the template is replaced by an unclosed \texttt{<think>} prefix, and the generation budget includes both reasoning and the answer. Complete templates and prompt variants are available as open source on \href{https://github.com/SonnyNondegeneracy/partition-function-llm}{GitHub} \cite{song2026partitioncode}.

\subsection{Evaluation methods}\label{sec:evaluation}

\paragraph{Parsing and counting.}
Invalid outputs receive zero implementation reward, create no state and are not resampled. Reward and potential estimation retain total counts, whereas entropy and scale calibration renormalise over valid responses, preserving failure effects while comparing uncertainty among valid destinations. Plotted mean reward is the mean implementation reward minus 1, so invalid outputs contribute \(-1\) to this measure. Invalid rates in the T1 and T2 main experiments are 0.26\% and 0.18\%; valid-only renormalisation changes mean reward and \(r\) in each round by at most 0.0065 and \(2.3\times10^{-4}\), respectively. Parsing, counting and implementation are open source on \href{https://github.com/SonnyNondegeneracy/partition-function-llm}{GitHub} \cite{song2026partitioncode}.

\paragraph{Pseudocounts in potential and order-parameter estimation.}
For the potential and the log ratios, use \(\widehat T_{ij}=(N_{ij}+\alpha)/(N_i+\alpha|\mathcal L(i)|)\), where \(N_i\) is the total sample count and \(N_{ij}\) the valid transition count; add \(\alpha=0.1\) pseudocounts to avoid divergent log ratios. The calculation of \(r\) uses only state pairs observed in both directions in the raw counts. For \(n\) states, the potential is estimated as in ref.~\cite{ref7}:
\begin{equation}\label{eq:potential-estimator}
\widehat\phi=\underset{\sum_i\phi_i=0}{\arg\min}
\frac1n\sum_i\sum_{j\in\mathcal L(i)}\widehat T_{ij}
e^{-(\phi_i-\phi_j)/2}
\end{equation}

\paragraph{Effective-scale calibration.}
For calibration, we add 0.5 pseudocounts to each valid destination and renormalise over valid destinations to prevent divergent log ratios. The same smoothed distribution is used to calculate the Jeffreys divergence and the mean reward conditional on valid responses, denoted by \(R^L_{im}\) in the figure.
We fit equation~\eqref{eq:calibration-proportional} through the origin by errors-in-variables regression, jointly estimating the calibration factor \(c_t>0\) and intrinsic scatter \cite{ref28}; measurement covariances between axes and round pairs within a source state come from joint count bootstrap resampling and enter the regression. For the main experiment, \(c_t\) is fitted jointly on the round pairs of the three training seeds.

\paragraph{Bootstrap intervals.}
All error bars are pointwise 95\% intervals from 500 bootstrap resamples (2.5\%--97.5\% quantiles) \cite{efron1993bootstrap}, with counts resampled according to each experiment's sampling design. These intervals reflect measurement uncertainty conditional on the fixed training trajectory. For the main experiments, Figs~\ref{fig:potential-emergence}--\ref{fig:training-ablations} also include retraining variability: the half-width is \(1.96(\sigma_{\mathrm{boot}}^2+\sigma_{\mathrm{seed}}^2)^{1/2}\), where \(\sigma_{\mathrm{boot}}\) is the bootstrap interval width divided by 3.92 and \(\sigma_{\mathrm{seed}}\) is the standard deviation over three training seeds for the same quantity.

\paragraph{Fitting theoretical curves.}
Theoretical curves use the \(\beta\to0\) asymptotic forms. We fit all 8 rounds by minimising the weighted residual sum \(\sum_m w_{X,m}[f_X(\beta_{\mathrm{eff},m})-X_m^{\mathrm{obs}}]^2\), with theoretical-remainder weights that give greater weight to small-\(\beta\) rounds. For both T1 and T2, order-parameter fits include the starting checkpoint and rounds 1--8, use \(\beta_{\mathrm{fit},m,t}^{-1}=\beta_{\mathrm{eff},m}^{-1}+\beta_{i,t}^{-1}\) (with \(\beta_{\mathrm{fit},0,t}=\beta_{i,t}\)), fix \(\kappa=0\), jointly fit \(\theta_t,\beta_{i,t}\), and use the weight scale \(s_r^2=\beta_{\mathrm{fit}}^6\). Specifically, the reward curve, theoretical remainder and weighting scale are \(f_R(\beta)=R_{\mathrm{max}}-Ae^{-\Delta/\beta}\), \(O(e^{-\delta/\beta})\) and \(s_R^2=e^{-4\Delta/\beta}\), respectively; for entropy they are \(f_H(\beta)=H_*+(A/\beta)e^{-\Delta/\beta}\), \(O(e^{-\Delta/\beta})\) and \(s_H^2=e^{-2\Delta/\beta}\); for the order parameter they are \(f_r(\beta)=(1+\kappa+\theta\beta^2)^{-1/2}\), \(O(\beta^3)\) and \(s_r^2=\beta^6\).

The general reward remainder involves \(\delta=\min(2\Delta,\Delta_2)\); its weights retain the scale \(s_R=e^{-2\Delta/\beta}\) from the two-term reward approximation above.
In fitting, we first divide the remainder scale by its maximum across rounds, then add a smoothing floor of \(0.30^2\) and normalise each observable's weights to a mean of 1:
\begin{equation*}
\begin{aligned}
v_{X,m}&=\left[\left(\frac{s_X(\beta_{\mathrm{eff},m})}{\max_\ell s_X(\beta_{\mathrm{eff},\ell})}\right)^2+0.30^2\right]^{-1},\\
w_{X,m}&=\frac{v_{X,m}}{\frac18\sum_{\ell=1}^{8}v_{X,\ell}}.
\end{aligned}
\end{equation*}

The three descriptive bands in Fig.~\ref{fig:reward-entropy} show the full theoretical-remainder scales \(Ae^{-2\Delta/\beta_{\mathrm{eff}}}\) for reward and \(Ae^{-\Delta/\beta_{\mathrm{eff}}}\) for entropy, using each series' fitted \(A,\Delta\). The \(t5\) fit fixes task-defined \(R_{\mathrm{max},t5},\Delta_{t5},H_{*,t5}\) and estimates only \(A_{t5}\); the T1 and T2 means jointly fit \(R_{\mathrm{max}},H_*,A,\Delta\). These descriptive bands contain no measurement uncertainty and are not confidence intervals.

Main experiments use calibrated \(\beta_{\mathrm{eff},m}\): for a single state, fix \(R_{\mathrm{max},i},\Delta_i,H_{*,i}\) and fit only \(A_i\); for task-level values, jointly fit \(R_{\mathrm{max}},H_*,A,\Delta\), updating weights for each candidate \(\Delta\). For main-experiment \(r\), fix \(\kappa=0\) and jointly fit \(\theta_t,\beta_{i,t}\) using the task-specific initial-scale adjustment above. Figure~\ref{fig:ablations} uses nominal \(\beta_{\mathrm{nominal},m}\).

\section{Data availability}
The data supporting the conclusions of this study are archived on Hugging Face \cite{song2026partitiondata}: \url{https://huggingface.co/datasets/Nondegeneracy/partition-function-llm}. The archive includes raw model responses, state-transition counts, statistical analysis results, cumulative LoRA adapter weights, experimental configurations and metadata, and source data for Figs.~2--5. The public source and fixed revision of the base model are recorded in the metadata. Original content in the archive is licensed under CC BY 4.0; third-party materials retain their original licences.

\section{Code availability}\label{sec:code-availability}
Code for model post-training, testing and measurement, data analysis and plotting is available as open source on GitHub \cite{song2026partitioncode}: \url{https://github.com/SonnyNondegeneracy/partition-function-llm}. The repository provides dependencies, instructions and reproducibility examples. Original code is licensed under the MIT licence; third-party code retains its original licences.

\section{Acknowledgements}
This work is supported by National Natural Science Foundation of China under contract No. 12425505. AI tools were used to assist with code development and language editing; the authors take full responsibility for the content of this paper.

\section{Competing interests}
The authors declare no competing interests.

\clearpage
\onecolumn
\section{Figure legends}
\noindent\textbf{Fig.~\ref{fig:task-design-theory} |} Agent task design and illustration of potential emergence during post-training. \textbf{a}, A safety-testing agent designs a test set, T1 selects the next test, and the selected test is then executed; T1 contains 8 states, with transitions allowed between any two distinct states. \textbf{b}, A physics problem-solving agent designs a sequence of solution steps, T2 selects the next step, and the corresponding operation is then executed; T2 contains 12 states, each connected to its two neighbours in either direction on a ring. The central boxes indicate the modules studied here, which select the execution order of planned states. \textbf{c}, Schematic of potential emergence under \(T_{ij}=Q_{ij}e^{R_{ij}/\beta}/Z_i(\beta)\) for five states. Bottom, order parameter \(r\) versus post-training progress \(-\log\beta\) (\(\beta\) is the KL temperature). Top, net flow (arrows) at the three marked points: as \(\beta\) decreases, a potential (node height) increasingly sets the direction of the net flow, which changes from a loop flow on a flat landscape to a flow down the potential.\par\medskip
\noindent\textbf{Fig.~\ref{fig:potential-emergence} |} Strengthening of potential structure during post-training and tests of its theoretical description. \textbf{a}, Change of each transition probability from the initial kernel, \(T_{ij}-Q_{ij}\): prediction of equation~\eqref{eq:optimal-kernel} at the calibrated \(\beta_{\mathrm{eff},m}\) versus measurement, for all transitions of T1 and T2. Rounds 1, 4 and 8 of the main experiment are shown; the Pearson correlations use all 8 rounds. Shading marks agreement in sign, and the dashed line denotes equality. \textbf{b}, Order parameter \(r\) for T1 and T2 as a function of calibrated \(\beta_{\mathrm{eff}}\). Solid lines show fits of equation~\eqref{eq:r-asymptotic} with \(\kappa=0\). Open markers denote measurements; error bars in b and c combine the 95\% interval from 500 bootstrap resamples with the spread over three training seeds (Section~\ref{sec:evaluation}). \textbf{c}, Centred measured potential \(\beta_{\mathrm{eff},8}\widehat\phi_i\) in round 8 versus the centred theoretical limit \(\Phi_i(0)=-u_i-\max_{k:Q_{ik}>0}u_k\). Each point represents one state. The dashed line denotes the theoretical equality \(\beta_{\mathrm{eff},8}\widehat\phi_i=\Phi_i(0)\). The Pearson correlations between the centred theoretical limit and centred measured potential are 0.86 for T1 and 0.61 for T2. The value of \(\beta_{\mathrm{eff},8}\) is taken from the calibration in Fig.~\ref{fig:reward-entropy}. We draw 4,096 samples per source per round. Panel b includes all 8 rounds.\par\medskip
\noindent\textbf{Fig.~\ref{fig:reward-entropy} |} Consistency tests of mean reward and state Shannon entropy during post-training. \textbf{a}, Proportional calibration of the effective scale. The horizontal axis shows the Jeffreys divergence between distributions from two rounds for the same source state; the vertical axis shows the corresponding reward difference multiplied by the difference in nominal inverse \(\beta\). Solid lines are fits through the origin. All 28 round pairs for each of the 8 T1 and 12 T2 source states are included, with circles and triangles distinguishing the tasks. \textbf{b,e}, Representative T1 source state \(t5\). \textbf{c,f}, Averages over the 8 T1 source states. \textbf{d,g}, Averages over the 12 T2 source states. The top row shows mean reward and the bottom row shows state Shannon entropy. Open markers denote measurements; solid lines evaluate the theoretical expressions at the calibrated \(\beta_{\mathrm{eff},m}=c_t\beta_{\mathrm{nominal},m}\). In b,e, \(R_{\mathrm{max},i}\), \(\Delta_i\) and \(H_{*,i}\) are determined by the predefined task, and only \(A_i\) is fitted with \(c_t\) fixed. In c,f and d,g, the parameters \(R_{\mathrm{max}},H_*,A,\Delta\) are jointly fitted for each task with its calibrated \(c_t\) fixed. Shading denotes theoretical remainder bands; their construction is described in Section~\ref{sec:evaluation}. Error bars in all panels combine the 95\% paired count bootstrap interval with the spread over three training seeds (Section~\ref{sec:evaluation}). We draw 4,096 samples per source per round.\par\medskip
\noindent\textbf{Fig.~\ref{fig:ablations} |} Sensitivity of post-training organisation and reward ablation. The five columns compare inference mode and sampling temperature, model size and family, prompt variants, evaluation of the same T1-trained checkpoints on T1 and T2, and reward structure, \(R_{ij}=u_j+\alpha\epsilon_{ij}\) with shared destination rewards (\(\alpha=0\), main experiment) and an added edge-specific residual (\(\alpha>0\), diamonds). \textbf{a--e}, Order parameter \(r\); \textbf{f--j}, mean reward \(R\); \textbf{k--o}, state Shannon entropy \(H\). All 8 rounds are plotted against \(-\log\beta_{\mathrm{nominal},m}\), where \(\beta_{\mathrm{nominal},m}=0.375/m\). Open markers denote measurements; error bars show 95\% interval of 500 bootstrap resamples. Some error bars are smaller than the markers. For the main T1 protocol (blue circles), error bars also include the spread over three training seeds (Section~\ref{sec:evaluation}). For prompt variants, statistics are computed per resample and averaged equally across variants before forming the intervals. Direct-answer and thinking-mode evaluations use 4,096 and 256 samples per source, respectively.\par\medskip
\noindent\textbf{Fig.~\ref{fig:training-ablations} |} Sensitivity to the training procedure in T1. \textbf{a}, Calibration relation of Fig.~\ref{fig:reward-entropy}a in effective units: \(\Delta(1/\beta_{\mathrm{eff}})\,\Delta R\) versus the Jeffreys divergence between two rounds, averaged over the 8 source states, for every round pair of the 8 branches. Each branch uses its own calibration factor \(c\); the dashed line denotes equality. \textbf{b}, Order parameter \(r\) of all other branches at their own calibrated \(\beta_{\mathrm{eff},m}\) (circles), with the main-experiment measurements and fit of Fig.~\ref{fig:potential-emergence}b (squares, line); no parameter is refitted. \textbf{c}, Calibration factor \(c\) versus the KL coefficient \(\beta_0\); the main experiment is the square, and the horizontal line denotes \(c=1\), the ideal optimum in every round. In b, error bars of the main experiment combine the 95\% interval from 500 bootstrap resamples with the spread over three training seeds; in c, the error bar shows the spread of \(c\) over the three seeds (Section~\ref{sec:evaluation}). Branches comprise the main experiment and seven further KL coefficients (Section~\ref{sec:experiments}); each branch is calibrated from all of its measured rounds. We draw 4,096 samples per source per round.\par\medskip
\clearpage
\twocolumn
\begingroup\small
\setlength{\bibsep}{5pt}
\bibliographystyle{naturemag}
\bibliography{main}

@article{ref23,
  title = {Scientific discovery in the age of artificial intelligence},
  volume = {620},
  issn = {1476-4687},
  url = {http://dx.doi.org/10.1038/s41586-023-06221-2},
  doi = {10.1038/s41586-023-06221-2},
  number = {7972},
  journal = {Nature},
  publisher = {Springer Science and Business Media LLC},
  author = {Wang, Hanchen and Fu, Tianfan and Du, Yuanqi and Gao, Wenhao and Huang, Kexin and Liu, Ziming and Chandak, Payal and Liu, Shengchao and Van Katwyk, Peter and Deac, Andreea and Anandkumar, Anima and Bergen, Karianne and Gomes, Carla P. and Ho, Shirley and Kohli, Pushmeet and Lasenby, Joan and Leskovec, Jure and Liu, Tie-Yan and Manrai, Arjun and Marks, Debora and Ramsundar, Bharath and Song, Le and Sun, Jimeng and Tang, Jian and Veličković, Petar and Welling, Max and Zhang, Linfeng and Coley, Connor W. and Bengio, Yoshua and Zitnik, Marinka},
  year = {2023},
  month = {Aug},
  pages = {47–60}
}

@article{ref24,
  title = {Autonomous chemical research with large language models},
  volume = {624},
  issn = {1476-4687},
  url = {http://dx.doi.org/10.1038/s41586-023-06792-0},
  doi = {10.1038/s41586-023-06792-0},
  number = {7992},
  journal = {Nature},
  publisher = {Springer Science and Business Media LLC},
  author = {Boiko, Daniil A. and MacKnight, Robert and Kline, Ben and Gomes, Gabe},
  year = {2023},
  month = {Dec},
  pages = {570–578}
}

@article{bran2023chemcrow,
  title = {Augmenting large language models with chemistry tools},
  volume = {6},
  issn = {2522-5839},
  url = {http://dx.doi.org/10.1038/s42256-024-00832-8},
  doi = {10.1038/s42256-024-00832-8},
  number = {5},
  journal = {Nature Machine Intelligence},
  publisher = {Springer Science and Business Media LLC},
  author = {M. Bran, Andres and Cox, Sam and Schilter, Oliver and Baldassari, Carlo and White, Andrew D. and Schwaller, Philippe},
  year = {2024},
  month = {May},
  pages = {525–535}
}

@article{funsearch,
  author  = {Romera-Paredes, Bernardino and Barekatain, Mohammadamin and Novikov, Alexander and Balog, Matej and Kumar, M. Pawan and Dupont, Emilien and Ruiz, Francisco J. R. and Ellenberg, Jordan S. and Wang, Pengming and Fawzi, Omar and Kohli, Pushmeet and Fawzi, Alhussein},
  title   = {Mathematical discoveries from program search with large language models},
  journal = {Nature},
  year    = {2024},
  volume  = {625},
  number  = {7995},
  pages   = {468--475},
  doi     = {10.1038/s41586-023-06924-6},
  url     = {https://doi.org/10.1038/s41586-023-06924-6},
  issn    = {1476-4687}
}

@misc{alphaevolve,
  title = {AlphaEvolve: A coding agent for scientific and algorithmic discovery},
  author = {Alexander Novikov and Ngân Vũ and Marvin Eisenberger and Emilien Dupont and Po-Sen Huang and Adam Zsolt Wagner and Sergey Shirobokov and Borislav Kozlovskii and Francisco J. R. Ruiz and Abbas Mehrabian and M. Pawan Kumar and Abigail See and Swarat Chaudhuri and George Holland and Alex Davies and Sebastian Nowozin and Pushmeet Kohli and Matej Balog},
  year = {2025},
  eprint = {2506.13131},
  archivePrefix = {arXiv},
  primaryClass = {cs.AI},
  url = {https://arxiv.org/abs/2506.13131}
}

@article{Song:2025pwy,
  title = {Explainable AI-assisted optimization for Feynman integral reduction},
  volume = {2026},
  issn = {1029-8479},
  url = {http://dx.doi.org/10.1007/jhep06(2026)225},
  doi = {10.1007/jhep06(2026)225},
  number = {6},
  journal = {Journal of High Energy Physics},
  publisher = {Springer Science and Business Media LLC},
  author = {Song, Zhuo-Yang and Yang, Tong-Zhi and Cao, Qing-Hong and Luo, Ming-xing and Zhu, Hua Xing},
  year = {2026},
  month = {June}
}

@misc{Cao:2025shc,
      title={Scalable Quantum State Preparation via Large-Language-Model-Driven Discovery}, 
      author={Qing-Hong Cao and Zong-Yue Hou and Ying-Ying Li and Xiaohui Liu and Zhuo-Yang Song and Liang-Qi Zhang and Shutao Zhang and Ke Zhao},
      year={2025},
      eprint={2505.06347},
      archivePrefix={arXiv},
      primaryClass={quant-ph},
      url={https://arxiv.org/abs/2505.06347}, 
}

@article{song2025iteratedagentsymbolicregression,
  title = {Iterated Agent for Symbolic Regression},
  issn = {1572-9494},
  url = {http://dx.doi.org/10.1088/1572-9494/aea105},
  doi = {10.1088/1572-9494/aea105},
  journal = {Communications in Theoretical Physics},
  publisher = {IOP Publishing},
  author = {Song, Zhuo-Yang and Cai, Zeyu and Zhang, Shu-Tao and Wei, Jiashen and Pan, Jichen and Qiu, Shi and Cao, Qing-Hong and Hou, Tie-Jiun and Liu, Xiaohui and Luo, Ming-xing and Zhu, Huaxing},
  year = {2026},
  month = {Sept}
}

@inproceedings{Mohammadi_2025,
  series = {KDD ’25},
  title = {Evaluation and Benchmarking of LLM Agents: A Survey},
  url = {http://dx.doi.org/10.1145/3711896.3736570},
  doi = {10.1145/3711896.3736570},
  booktitle = {Proceedings of the 31st ACM SIGKDD Conference on Knowledge Discovery and Data Mining V.2},
  publisher = {ACM},
  author = {Mohammadi, Mahmoud and Li, Yipeng and Lo, Jane and Yip, Wendy},
  year = {2025},
  month = {Aug},
  pages = {6129–6139},
  collection = {KDD ’25}
}

@article{sapkota2025aiagentsvsagentic,
title = {AI Agents vs. Agentic AI: A Conceptual taxonomy, applications and challenges},
journal = {Information Fusion},
volume = {126},
pages = {103599},
year = {2026},
issn = {1566-2535},
doi = {https://doi.org/10.1016/j.inffus.2025.103599},
url = {https://www.sciencedirect.com/science/article/pii/S1566253525006712},
author = {Ranjan Sapkota and Konstantinos I. Roumeliotis and Manoj Karkee}
}

@inproceedings{yao2023reactsynergizingreasoningacting,
title={ReAct: Synergizing Reasoning and Acting in Language Models},
author={Shunyu Yao and Jeffrey Zhao and Dian Yu and Nan Du and Izhak Shafran and Karthik R Narasimhan and Yuan Cao},
booktitle={The Eleventh International Conference on Learning Representations },
year={2023},
url={https://openreview.net/forum?id=WE_vluYUL-X}
}

@inproceedings{schick2023toolformerlanguagemodelsteach,
  series = {NeurIPS 2023},
  title = {Toolformer: Language Models Can Teach Themselves to Use Tools},
  url = {http://dx.doi.org/10.52202/075280-2997},
  doi = {10.52202/075280-2997},
  booktitle = {Advances in Neural Information Processing Systems 36},
  publisher = {Neural Information Processing Systems Foundation, Inc. (NeurIPS)},
  author = {Schick, Timo and Dwivedi-Yu, Jane and Dessi, Roberto and Raileanu, Roberta and Lomeli, Maria and Hambro, Eric and Zettlemoyer, Luke and Cancedda, Nicola and Scialom, Thomas},
  year = {2023},
  pages = {68539–68551},
  collection = {NeurIPS 2023}
}

@article{wang2023voyageropenendedembodiedagent,
title={Voyager: An Open-Ended Embodied Agent with Large Language Models},
author={Guanzhi Wang and Yuqi Xie and Yunfan Jiang and Ajay Mandlekar and Chaowei Xiao and Yuke Zhu and Linxi Fan and Anima Anandkumar},
journal={Transactions on Machine Learning Research},
issn={2835-8856},
year={2024},
url={https://openreview.net/forum?id=ehfRiF0R3a},
note={}
}

@inproceedings{Park2023,
  series = {UIST ’23},
  title = {Generative Agents: Interactive Simulacra of Human Behavior},
  url = {http://dx.doi.org/10.1145/3586183.3606763},
  doi = {10.1145/3586183.3606763},
  booktitle = {Proceedings of the 36th Annual ACM Symposium on User Interface Software and Technology},
  publisher = {ACM},
  author = {Park, Joon Sung and O’Brien, Joseph and Cai, Carrie Jun and Morris, Meredith Ringel and Liang, Percy and Bernstein, Michael S.},
  year = {2023},
  month = {Oct},
  pages = {1–22},
  collection = {UIST ’23}
}

@inproceedings{NEURIPS2022_9d560961,
  series = {NeurIPS 2022},
  title = {Chain-Of-Thought Prompting Elicits Reasoning in Large Language Models},
  url = {http://dx.doi.org/10.52202/068431-1800},
  DOI = {10.52202/068431-1800},
  booktitle = {Advances in Neural Information Processing Systems 35},
  publisher = {Neural Information Processing Systems Foundation,  Inc. (NeurIPS)},
  author = {Wei,  Jason and Wang,  Xuezhi and Schuurmans,  Dale and Bosma,  Maarten and Ichter,  Brian and Xia,  Fei and Chi,  Ed and Le,  Quoc V and Zhou,  Denny},
  year = {2022},
  pages = {24824–24837},
  collection = {NeurIPS 2022}
}

@inproceedings{wang2023selfconsistencyimproveschainthought,
title={Self-Consistency Improves Chain of Thought Reasoning in Language Models},
author={Xuezhi Wang and Jason Wei and Dale Schuurmans and Quoc V Le and Ed H. Chi and Sharan Narang and Aakanksha Chowdhery and Denny Zhou},
booktitle={The Eleventh International Conference on Learning Representations },
year={2023},
url={https://openreview.net/forum?id=1PL1NIMMrw}
}

@inproceedings{wang2023planandsolvepromptingimprovingzeroshot,
      title={Plan-and-Solve Prompting: Improving Zero-Shot Chain-of-Thought Reasoning by Large Language Models},
      author={Lei Wang and Wanyu Xu and Yihuai Lan and Zhiqiang Hu and Yunshi Lan and Roy Ka-Wei Lee and Ee-Peng Lim},
      booktitle={Proceedings of the 61st Annual Meeting of the Association for Computational Linguistics (Volume 1: Long Papers)},
      pages={2609--2634},
      year={2023},
      doi={10.18653/v1/2023.acl-long.147},
      url={https://aclanthology.org/2023.acl-long.147},
}

@article{chen2023programthoughtspromptingdisentangling,
title={Program of Thoughts Prompting: Disentangling Computation from Reasoning for Numerical Reasoning Tasks},
author={Wenhu Chen and Xueguang Ma and Xinyi Wang and William W. Cohen},
journal={Transactions on Machine Learning Research},
issn={2835-8856},
year={2023},
url={https://openreview.net/forum?id=YfZ4ZPt8zd},
note={}
}

@InProceedings{pmlr-v202-gao23f,
  title = 	 {{PAL}: Program-aided Language Models},
  author =       {Gao, Luyu and Madaan, Aman and Zhou, Shuyan and Alon, Uri and Liu, Pengfei and Yang, Yiming and Callan, Jamie and Neubig, Graham},
  booktitle = 	 {Proceedings of the 40th International Conference on Machine Learning},
  pages = 	 {10764--10799},
  year = 	 {2023},
  editor = 	 {Krause, Andreas and Brunskill, Emma and Cho, Kyunghyun and Engelhardt, Barbara and Sabato, Sivan and Scarlett, Jonathan},
  volume = 	 {202},
  series = 	 {Proceedings of Machine Learning Research},
  month = 	 {23--29 Jul},
  publisher =    {PMLR},
  url = 	 {https://proceedings.mlr.press/v202/gao23f.html}
}

@inproceedings{yao2023treethoughtsdeliberateproblem,
  series = {NeurIPS 2023},
  title = {Tree of Thoughts: Deliberate Problem Solving with Large Language Models},
  url = {http://dx.doi.org/10.52202/075280-0517},
  doi = {10.52202/075280-0517},
  booktitle = {Advances in Neural Information Processing Systems 36},
  publisher = {Neural Information Processing Systems Foundation, Inc. (NeurIPS)},
  author = {Yao, Shunyu and Yu, Dian and Zhao, Jeffrey and Shafran, Izhak and Griffiths, Tom and Cao, Yuan and Narasimhan, Karthik},
  year = {2023},
  pages = {11809–11822},
  collection = {NeurIPS 2023}
}

@article{Besta_Blach_Kubicek_Gerstenberger_Podstawski_Gianinazzi_Gajda_Lehmann_Niewiadomski_Nyczyk_Hoefler_2024, 
   title={Graph of Thoughts: Solving Elaborate Problems with Large Language Models},
   volume={38},
   ISSN={2159-5399},
   url={http://dx.doi.org/10.1609/aaai.v38i16.29720},
   DOI={10.1609/aaai.v38i16.29720},
   number={16},
   journal={Proceedings of the AAAI Conference on Artificial Intelligence},
   publisher={Association for the Advancement of Artificial Intelligence (AAAI)},
   author={Besta, Maciej and Blach, Nils and Kubicek, Ales and Gerstenberger, Robert and Podstawski, Michal and Gianinazzi, Lukas and Gajda, Joanna and Lehmann, Tomasz and Niewiadomski, Hubert and Nyczyk, Piotr and Hoefler, Torsten},
   year={2024},
   month={Mar.}, pages={17682–17690} }

@inproceedings{liang2023codepolicieslanguagemodel,
  author={Liang, Jacky and Huang, Wenlong and Xia, Fei and Xu, Peng and Hausman, Karol and Ichter, Brian and Florence, Pete and Zeng, Andy},
  booktitle={2023 IEEE International Conference on Robotics and Automation (ICRA)}, 
  title={Code as Policies: Language Model Programs for Embodied Control}, 
  publisher={IEEE},
  address={Piscataway, NJ},
  year={2023},
  volume={},
  number={},
  pages={9493-9500},
  doi={10.1109/ICRA48891.2023.10160591},
  url = {https://doi.org/10.1109/icra48891.2023.10160591}
}

@misc{ref30,
      title={Fork-and-Flush: Escaping Idea Basins in Autoresearch Agents}, 
      author={Ziyang Cai and Christos Ziakas and Vasilis Kontonis and Tim Pearce and Siddhartha Sen and Akshay Krishnamurthy and Shivam Garg and Dimitris Papailiopoulos},
      year={2026},
      eprint={2610.07447},
      archivePrefix={arXiv},
      primaryClass={cs.LG},
      url={https://arxiv.org/abs/2610.07447}, 
}

@article{ref21,
  title = {Emergence of polarization in networks of large language model agents},
  issn = {2041-1723},
  url = {http://dx.doi.org/10.1038/s41467-026-78228-y},
  doi = {10.1038/s41467-026-78228-y},
  journal = {Nature Communications},
  publisher = {Springer Science and Business Media LLC},
  author = {Piao, Jinghua and Lu, Zhihong and Gao, Chen and Xu, Fengli and Hu, Qinghua and Santos, Fernando P. and Li, Yong and Evans, James},
  year = {2026},
  month = {Oct}
}

@misc{ref22,
      title={Superintelligent Agents Pose Catastrophic Risks: Can Scientist AI Offer a Safer Path?}, 
      author={Yoshua Bengio and Michael Cohen and Damiano Fornasiere and Joumana Ghosn and Pietro Greiner and Matt MacDermott and Sören Mindermann and Adam Oberman and Jesse Richardson and Oliver Richardson and Marc-Antoine Rondeau and Pierre-Luc St-Charles and David Williams-King},
      year={2025},
      eprint={2502.15657},
      archivePrefix={arXiv},
      primaryClass={cs.AI},
      url={https://arxiv.org/abs/2502.15657}, 
}

@article{kim2025sciencescalingagentsystems,
  title = {Capable language models can outgrow the benefits of collaboration},
  volume = {8},
  issn = {2522-5839},
  url = {http://dx.doi.org/10.1038/s42256-026-01268-y},
  doi = {10.1038/s42256-026-01268-y},
  number = {7},
  journal = {Nature Machine Intelligence},
  publisher = {Springer Science and Business Media LLC},
  author = {Kim, Yubin and Gu, Ken and Park, Chanwoo and Park, Chunjong and Schmidgall, Samuel and Heydari, A. Ali and Yan, Yao and Zhang, Zhihan and Zhuang, Yuchen and Liu, Yun and Malhotra, Mark and Liang, Paul Pu and Won Park, Hae and Yang, Yuzhe and Xu, Xuhai and Du, Yilun and Patel, Shwetak and Althoff, Tim and McDuff, Daniel and Liu, Xin},
  year = {2026},
  month = {July},
  pages = {1157–1172}
}

@misc{liu2025neuralthermodynamiclawslarge,
      title={Neural Thermodynamic Laws for Large Language Model Training}, 
      author={Ziming Liu and Yizhou Liu and Jeff Gore and Max Tegmark},
      year={2025},
      eprint={2505.10559},
      archivePrefix={arXiv},
      primaryClass={cs.LG},
      url={https://arxiv.org/abs/2505.10559}, 
}

@inproceedings{ref1,
title={Semantic Uncertainty: Linguistic Invariances for Uncertainty Estimation in Natural Language Generation},
author={Lorenz Kuhn and Yarin Gal and Sebastian Farquhar},
booktitle={The Eleventh International Conference on Learning Representations },
year={2023},
url={https://openreview.net/forum?id=VD-AYtP0dve}
}

@article{ref2,
  title = {Detecting hallucinations in large language models using semantic entropy},
  volume = {630},
  issn = {1476-4687},
  url = {http://dx.doi.org/10.1038/s41586-024-07421-0},
  doi = {10.1038/s41586-024-07421-0},
  number = {8017},
  journal = {Nature},
  publisher = {Springer Science and Business Media LLC},
  author = {Farquhar, Sebastian and Kossen, Jannik and Kuhn, Lorenz and Gal, Yarin},
  year = {2024},
  month = {June},
  pages = {625–630}
}

@inproceedings{ref20,
  title = 	 {Metastable Dynamics of Chain-of-Thought Reasoning: Provable Benefits of Search, {RL} and Distillation},
  author =       {Kim, Juno and Wu, Denny and Lee, Jason D. and Suzuki, Taiji},
  booktitle = 	 {Proceedings of the 42nd International Conference on Machine Learning},
  pages = 	 {30791--30825},
  year = 	 {2025},
  editor = 	 {Singh, Aarti and Fazel, Maryam and Hsu, Daniel and Lacoste-Julien, Simon and Berkenkamp, Felix and Maharaj, Tegan and Wagstaff, Kiri and Zhu, Jerry},
  volume = 	 {267},
  series = 	 {Proceedings of Machine Learning Research},
  month = 	 {13--19 Jul},
  publisher =    {PMLR},
  url = 	 {https://proceedings.mlr.press/v267/kim25aj.html}
}

@misc{ref7,
      title={Detailed balance in large language model-driven agents}, 
      author={Zhuo-Yang Song and Qing-Hong Cao and Ming-xing Luo and Hua Xing Zhu},
      journal={PRX Intelligence},
      note={Accepted as a Research Article},
      year={2026},
      eprint={2512.10047},
      archivePrefix={arXiv},
      primaryClass={cs.LG},
      doi={10.1103/kwx6-clqd},
      url={https://doi.org/10.1103/kwx6-clqd},
}

@inproceedings{ref3,
 author = {Kirk, Robert and Mediratta, Ishita and Nalmpantis, Christoforos and Luketina, Jelena and Hambro, Eric and Grefenstette, Edward and Raileanu, Roberta},
 booktitle = {International Conference on Learning Representations},
 editor = {B. Kim and Y. Yue and S. Chaudhuri and K. Fragkiadaki and M. Khan and Y. Sun},
 pages = {20620--20653},
 title = {Understanding the Effects of RLHF on LLM Generalisation and Diversity},
 url = {https://proceedings.iclr.cc/paper_files/paper/2024/file/5a68d05006d5b05dd9463dd9c0219db0-Paper-Conference.pdf},
 volume = {2024},
 year = {2024}
}

@article{ref4,
  title = {On the Algorithmic Bias of Aligning Large Language Models with RLHF: Preference Collapse and Matching Regularization},
  volume = {120},
  issn = {1537-274X},
  url = {http://dx.doi.org/10.1080/01621459.2025.2555067},
  doi = {10.1080/01621459.2025.2555067},
  number = {552},
  journal = {Journal of the American Statistical Association},
  publisher = {Informa UK Limited},
  author = {Xiao, Jiancong and Li, Ziniu and Xie, Xingyu and Getzen, Emily and Fang, Cong and Long, Qi and Su, Weijie J.},
  year = {2025},
  month = {Oct},
  pages = {2154–2164}
}

@inproceedings{ref26,
    series = {NeurIPS 2025},
  title = {Does Reinforcement Learning Really Incentivize Reasoning Capacity in LLMs Beyond the Base Model?},
  url = {http://dx.doi.org/10.52202/085713-1933},
  DOI = {10.52202/085713-1933},
  booktitle = {Advances in Neural Information Processing Systems 38},
  publisher = {Neural Information Processing Systems Foundation,  Inc. (NeurIPS)},
  author = {Chen,  Zhiqi and Lu,  Rui and Zhao,  Andrew and Wang,  Zhaokai and Yue,  Yang and Song,  Shiji and Huang,  Gao},
  year = {2025},
  pages = {64304–64339},
  collection = {NeurIPS 2025}
}

@misc{ref5,
      title={From Simulation to Enaction: Post-trained language models recognize and react to their own generations}, 
      author={Asvin G. and Jack Lindsey},
      year={2026},
      eprint={2605.25459},
      archivePrefix={arXiv},
      primaryClass={cs.LG},
      url={https://arxiv.org/abs/2605.25459}, 
}

@misc{ref6,
      title={The Assistant as a Privileged Persona: A canonical reference in cross-persona self-recognition}, 
      author={Asvin G},
      year={2026},
      eprint={2606.00545},
      archivePrefix={arXiv},
      primaryClass={cs.LG},
      url={https://arxiv.org/abs/2606.00545}, 
}

@misc{hu2025conceptnetwork,
      title={Emergent Slow Thinking in {LLMs} as Inverse Tree Freezing},
      author={Sihan Hu and Xiansheng Cai and Yuan Huang and Zhiyuan Yao and Linfeng Zhang and Pan Zhang and Youjin Deng and Kun Chen},
      year={2026},
      eprint={2509.23629},
      archivePrefix={arXiv},
      primaryClass={cs.AI},
      url={https://arxiv.org/abs/2509.23629}, 
}

@article{ref9,
  title = {Linear Theory for Control of Nonlinear Stochastic Systems},
  volume = {95},
  issn = {1079-7114},
  url = {http://dx.doi.org/10.1103/physrevlett.95.200201},
  doi = {10.1103/physrevlett.95.200201},
  number = {20},
  journal = {Physical Review Letters},
  publisher = {American Physical Society (APS)},
  author = {Kappen, Hilbert J.},
  year = {2005},
  month = {Nov}
}

@article{ref10,
  title = {Efficient computation of optimal actions},
  volume = {106},
  issn = {1091-6490},
  url = {http://dx.doi.org/10.1073/pnas.0710743106},
  doi = {10.1073/pnas.0710743106},
  number = {28},
  journal = {Proceedings of the National Academy of Sciences},
  publisher = {National Academy of Sciences},
  author = {Todorov, Emanuel},
  year = {2009},
  month = {July},
  pages = {11478–11483}
}

@inproceedings{ref11,
  title = 	 {Reinforcement Learning with Deep Energy-Based Policies},
  author =       {Tuomas Haarnoja and Haoran Tang and Pieter Abbeel and Sergey Levine},
  booktitle = 	 {Proceedings of the 34th International Conference on Machine Learning},
  pages = 	 {1352--1361},
  year = 	 {2017},
  editor = 	 {Precup, Doina and Teh, Yee Whye},
  volume = 	 {70},
  series = 	 {Proceedings of Machine Learning Research},
  month = 	 {06--11 Aug},
  publisher =    {PMLR},
  url = 	 {https://proceedings.mlr.press/v70/haarnoja17a.html}
}

@misc{ref19,
  title = {Reinforcement Learning and Control as Probabilistic Inference: Tutorial and Review},
  author = {Sergey Levine},
  year = {2018},
  eprint = {1805.00909},
  archivePrefix = {arXiv},
  primaryClass = {cs.LG},
  url = {https://arxiv.org/abs/1805.00909}
}

@misc{qwen35_2026,
  title = {{Qwen3.5}: Towards Native Multimodal Agents},
  author = {{Qwen Team}},
  year = {2026},
  month = feb,
  url = {https://qwen.ai/blog?id=qwen3.5}
}

@inproceedings{hu2022lora,
title={Lo{RA}: Low-Rank Adaptation of Large Language Models},
author={Edward J Hu and yelong shen and Phillip Wallis and Zeyuan Allen-Zhu and Yuanzhi Li and Shean Wang and Lu Wang and Weizhu Chen},
booktitle={International Conference on Learning Representations},
year={2022},
url={https://openreview.net/forum?id=nZeVKeeFYf9}
}

@article{williams1992reinforce,
  title = {Simple statistical gradient-following algorithms for connectionist reinforcement learning},
  volume = {8},
  issn = {1573-0565},
  url = {http://dx.doi.org/10.1007/bf00992696},
  doi = {10.1007/bf00992696},
  number = {3-4},
  journal = {Machine Learning},
  publisher = {Springer Science and Business Media LLC},
  author = {Williams, Ronald J.},
  year = {1992},
  month = {May},
  pages = {229–256}
}

@article{ref27,
title={Emergent Abilities of Large Language Models},
author={Jason Wei and Yi Tay and Rishi Bommasani and Colin Raffel and Barret Zoph and Sebastian Borgeaud and Dani Yogatama and Maarten Bosma and Denny Zhou and Donald Metzler and Ed H. Chi and Tatsunori Hashimoto and Oriol Vinyals and Percy Liang and Jeff Dean and William Fedus},
journal={Transactions on Machine Learning Research},
issn={2835-8856},
year={2022},
url={https://openreview.net/forum?id=yzkSU5zdwD},
note={Survey Certification}
}

@book{cover2006information,
  title = {Elements of Information Theory},
  isbn = {9780471748823},
  url = {http://dx.doi.org/10.1002/047174882x},
  doi = {10.1002/047174882x},
  publisher = {Wiley},
  author = {Cover, Thomas M. and Thomas, Joy A.},
  year = {2005},
  month = {Apr}
}

@misc{ref14,
  title = {The Entropy Mechanism of Reinforcement Learning for Reasoning Language Models},
  author = {Ganqu Cui and Yuchen Zhang and Jiacheng Chen and Lifan Yuan and Zhi Wang and Yuxin Zuo and Haozhan Li and Yuchen Fan and Huayu Chen and Weize Chen and Zhiyuan Liu and Hao Peng and Lei Bai and Wanli Ouyang and Yu Cheng and Bowen Zhou and Ning Ding},
  year = {2025},
  eprint = {2505.22617},
  archivePrefix = {arXiv},
  primaryClass = {cs.LG},
  url = {https://arxiv.org/abs/2505.22617}
}

@misc{ref8,
  title = {A Theoretical Lens for RL-Tuned Language Models via Energy-Based Models},
  author = {Zhiquan Tan and Yinrong Hong},
  year = {2025},
  eprint = {2512.18730},
  archivePrefix = {arXiv},
  primaryClass = {cs.LG},
  url = {https://arxiv.org/abs/2512.18730}
}

@inproceedings{ref12,
  series = {NeurIPS 2025},
  title = {Tracing the Representation Geometry of Language Models from Pretraining to Post-training},
  url = {http://dx.doi.org/10.52202/085713-1829},
  DOI = {10.52202/085713-1829},
  booktitle = {Advances in Neural Information Processing Systems 38},
  publisher = {Neural Information Processing Systems Foundation,  Inc. (NeurIPS)},
  author = {Li,  Melody and Agrawal,  Kumar Krishna and Ghosh,  Arna and Teru,  Komal and Santoro,  Adam and Lajoie,  Guillaume and Richards,  Blake},
  year = {2025},
  pages = {60929–60962},
  collection = {NeurIPS 2025}
}

@misc{ref13,
  title = {Manifold Steering Reveals the Shared Geometry of Neural Network Representation and Behavior},
  author = {Daniel Wurgaft and Can Rager and Matthew Kowal and Vasudev Shyam and Sheridan Feucht and Usha Bhalla and Tal Haklay and Eric Bigelow and Raphael Sarfati and Thomas McGrath and Owen Lewis and Jack Merullo and Noah Goodman and Thomas Fel and Atticus Geiger and Ekdeep Singh Lubana},
  year = {2026},
  eprint = {2605.05115},
  archivePrefix = {arXiv},
  primaryClass = {cs.LG},
  url = {https://arxiv.org/abs/2605.05115}
}

@misc{ref15,
  title = {Clip-Low Increases Entropy and Clip-High Decreases Entropy in Reinforcement Learning of Large Language Models},
  author = {Jaesung R. Park and Junsu Kim and Gyeongman Kim and Jinyoung Jo and Sean Choi and Jaewoong Cho and Ernest K. Ryu},
  year = {2025},
  eprint = {2509.26114},
  archivePrefix = {arXiv},
  primaryClass = {cs.LG},
  url = {https://arxiv.org/abs/2509.26114}
}

@article{Jiang2011graph,
  author    = {Jiang, Xiaoye and Lim, Lek-Heng and Yao, Yuan and Ye, Yinyu},
  title     = {Statistical ranking and combinatorial Hodge theory},
  journal   = {Mathematical Programming},
  volume    = {127},
  number    = {1},
  pages     = {203--244},
  year      = {2011},
  month     = {Mar.},
  issn      = {1436-4646},
  doi       = {10.1007/s10107-010-0419-x},
  url       = {https://doi.org/10.1007/s10107-010-0419-x}
}

@inproceedings{hu2025openrlhf,
    title = "{O}pen{RLHF}: A Ray-based Easy-to-use, Scalable and High-performance {RLHF} Framework",
    author = "Hu, Jian  and
      Wu, Xibin  and
      Shen, Wei  and
      Liu, Jason Klein  and
      Wang, Weixun  and
      Jiang, Songlin  and
      Wang, Haoran  and
      Chen, Hao  and
      Chen, Bin  and
      Fang, Wenkai  and
      Xianyu  and
      Cao, Yu  and
      Xu, Haotian  and
      Liu, Yiming",
    editor = {Habernal, Ivan  and
      Schulam, Peter  and
      Tiedemann, J{\"o}rg},
    booktitle = "Proceedings of the 2025 Conference on Empirical Methods in Natural Language Processing: System Demonstrations",
    month = nov,
    year = "2025",
    address = "Suzhou, China",
    publisher = "Association for Computational Linguistics",
    url = "https://aclanthology.org/2025.emnlp-demos.48/",
    doi = "10.18653/v1/2025.emnlp-demos.48",
    pages = "656--666",
    ISBN = "979-8-89176-334-0"
}

@misc{mangrulkar2022peft,
  title =        {{PEFT}: State-of-the-art Parameter-Efficient Fine-Tuning methods},
  author =       {Sourab Mangrulkar and Sylvain Gugger and Lysandre Debut and Younes Belkada and Sayak Paul and Benjamin Bossan and Marian Tietz},
  howpublished = {\url{https://github.com/huggingface/peft}},
  year =         {2022}
}

@inproceedings{kwon2023vllm,
  series = {SOSP ’23},
  title = {Efficient Memory Management for Large Language Model Serving with PagedAttention},
  url = {http://dx.doi.org/10.1145/3600006.3613165},
  doi = {10.1145/3600006.3613165},
  booktitle = {Proceedings of the 29th Symposium on Operating Systems Principles},
  publisher = {ACM},
  author = {Kwon, Woosuk and Li, Zhuohan and Zhuang, Siyuan and Sheng, Ying and Zheng, Lianmin and Yu, Cody Hao and Gonzalez, Joseph and Zhang, Hao and Stoica, Ion},
  year = {2023},
  month = {Oct},
  pages = {611–626},
  collection = {SOSP ’23}
}

@misc{song2026partitioncode,
  author = {Song, Zhuo-Yang and Zhu, Hua Xing},
  title = {A partition function theory of learning and inference in {LLM} agents: research code},
  year = {2026},
  howpublished = {GitHub},
  url = {https://github.com/SonnyNondegeneracy/partition-function-llm},
}

@misc{schulman2020kl,
  author = {Schulman, John},
  title = {Approximating {KL} Divergence},
  year = {2020},
  month = mar,
  howpublished = {Online technical note},
  url = {http://joschu.net/blog/kl-approx.html},
  note = {7 March 2020}
}

@inproceedings{kingma2015adam,
      title={Adam: A Method for Stochastic Optimization}, 
      author={Diederik P. Kingma and Jimmy Ba},
  booktitle = {International Conference on Learning Representations},
  year = {2015},
  url = {https://arxiv.org/abs/1412.6980}
}

@inproceedings{loshchilov2019adamw,
title={Decoupled Weight Decay Regularization},
author={Ilya Loshchilov and Frank Hutter},
booktitle={International Conference on Learning Representations},
year={2019},
url={https://openreview.net/forum?id=Bkg6RiCqY7},
}

@misc{schulman2017ppo,
  title = {Proximal Policy Optimization Algorithms},
  author = {John Schulman and Filip Wolski and Prafulla Dhariwal and Alec Radford and Oleg Klimov},
  year = {2017},
  eprint = {1707.06347},
  archivePrefix = {arXiv},
  primaryClass = {cs.LG},
  url = {https://arxiv.org/abs/1707.06347}
}

@misc{ref28,
      title={Data analysis recipes: Fitting a model to data}, 
      author={David W. Hogg and Jo Bovy and Dustin Lang},
      year={2010},
      eprint={1008.4686},
      archivePrefix={arXiv},
      primaryClass={astro-ph.IM},
      url={https://arxiv.org/abs/1008.4686}, 
}

@book{efron1993bootstrap,
  title = {An Introduction to the Bootstrap},
  isbn = {9781489945419},
  url = {http://dx.doi.org/10.1007/978-1-4899-4541-9},
  doi = {10.1007/978-1-4899-4541-9},
  publisher = {Springer US},
  author = {Efron, Bradley and Tibshirani, Robert J.},
  year = {1993}
}

@misc{song2026partitiondata,
  author = {Song, Zhuo-Yang and Zhu, Hua Xing},
  title = {A partition function theory of learning and inference in {LLM} agents: experimental data and adapters},
  year = {2026},
  howpublished = {Hugging Face},
  url = {https://huggingface.co/datasets/Nondegeneracy/partition-function-llm}
}

@article{Markov2006ExtensionOT,
author = {Markov, A.A.},
journal = {Journal Électronique d'Histoire des Probabilités et de la Statistique [electronic only]},
language = {eng},
number = {1b},
pages = {Article 10, 12 p., electronic only-Article 10, 12 p., electronic only},
publisher = {Séminaire d'Histoire du Calcul des Probabilités et de la Statistique, EHESS, Paris; Laboratoire de Probabilités et Modéles Aléatoires, Univerité Paris VI et VII},
title = {Extension of the law of large numbers to quantities, depending on each other (1906). Reprint.},
url = {http://eudml.org/doc/128778},
volume = {2},
year = {2006},
}

@misc{du2026reliability,
      title={Reliability-Safety Trade-off in AI Distillation: A Renormalization-Group Approach}, 
      author={Y. M. Du and Miao-Miao Yi and Tan-Ji Zhou and C. P. Sun},
      year={2026},
      eprint={2608.08572},
      archivePrefix={arXiv},
      primaryClass={cond-mat.stat-mech},
      url={https://arxiv.org/abs/2608.08572}, 
}

@inproceedings{vaswani2017attention,
 author = {Vaswani, Ashish and Shazeer, Noam and Parmar, Niki and Uszkoreit, Jakob and Jones, Llion and Gomez, Aidan N and Kaiser, \L ukasz and Polosukhin, Illia},
 booktitle = {Advances in Neural Information Processing Systems},
 editor = {I. Guyon and U. Von Luxburg and S. Bengio and H. Wallach and R. Fergus and S. Vishwanathan and R. Garnett},
 pages = {},
 publisher = {Curran Associates, Inc.},
 title = {Attention is All you Need},
 url = {https://proceedings.neurips.cc/paper_files/paper/2017/file/3f5ee243547dee91fbd053c1c4a845aa-Paper.pdf},
 volume = {30},
 year = {2017}
}

@inproceedings{brown2020language,
 author = {Brown, Tom and Mann, Benjamin and Ryder, Nick and Subbiah, Melanie and Kaplan, Jared D and Dhariwal, Prafulla and Neelakantan, Arvind and Shyam, Pranav and Sastry, Girish and Askell, Amanda and Agarwal, Sandhini and Herbert-Voss, Ariel and Krueger, Gretchen and Henighan, Tom and Child, Rewon and Ramesh, Aditya and Ziegler, Daniel and Wu, Jeffrey and Winter, Clemens and Hesse, Chris and Chen, Mark and Sigler, Eric and Litwin, Mateusz and Gray, Scott and Chess, Benjamin and Clark, Jack and Berner, Christopher and McCandlish, Sam and Radford, Alec and Sutskever, Ilya and Amodei, Dario},
 booktitle = {Advances in Neural Information Processing Systems},
 editor = {H. Larochelle and M. Ranzato and R. Hadsell and M.F. Balcan and H. Lin},
 pages = {1877--1901},
 publisher = {Curran Associates, Inc.},
 title = {Language Models are Few-Shot Learners},
 url = {https://proceedings.neurips.cc/paper_files/paper/2020/file/1457c0d6bfcb4967418bfb8ac142f64a-Paper.pdf},
 volume = {33},
 year = {2020}
}

@misc{kaplan2020scaling,
      title={Scaling Laws for Neural Language Models}, 
      author={Jared Kaplan and Sam McCandlish and Tom Henighan and Tom B. Brown and Benjamin Chess and Rewon Child and Scott Gray and Alec Radford and Jeffrey Wu and Dario Amodei},
      year={2020},
      eprint={2001.08361},
      archivePrefix={arXiv},
      primaryClass={cs.LG},
      url={https://arxiv.org/abs/2001.08361}, 
}

@inproceedings{hoffmann2022training, series={NeurIPS 2022}, title={Training Compute-Optimal Large Language Models}, url={http://dx.doi.org/10.52202/068431-2176}, DOI={10.52202/068431-2176}, booktitle={Advances in Neural Information Processing Systems 35}, publisher={Neural Information Processing Systems Foundation, Inc. (NeurIPS)}, author={Hoffmann, Jordan and Borgeaud, Sebastian and Mensch, Arthur and Buchatskaya, Elena and Cai, Trevor and Rutherford, Eliza and Casas, Diego De Las and Hendricks, Lisa Anne and Welbl, Johannes and Clark, Aidan and Hennigan, Thomas and Noland, Eric and Millican, Katherine and Driessche, George Van Den and Damoc, Bogdan and Guy, Aurelia and Osindero, Simon and Simonyan, Karén and Elsen, Erich and Vinyals, Oriol and Rae, Jack and Sifre, Laurent}, year={2022}, pages={30016–30030}, collection={NeurIPS 2022} }

@misc{lu2024aiscientist,
      title={The {AI} Scientist: Towards Fully Automated Open-Ended Scientific Discovery}, 
      author={Chris Lu and Cong Lu and Robert Tjarko Lange and Jakob Foerster and Jeff Clune and David Ha},
      year={2024},
      eprint={2408.06292},
      archivePrefix={arXiv},
      primaryClass={cs.AI},
      url={https://arxiv.org/abs/2408.06292}, 
}

@misc{xi2023rise,
      title={The Rise and Potential of Large Language Model Based Agents: A Survey}, 
      author={Zhiheng Xi and Wenxiang Chen and Xin Guo and Wei He and Yiwen Ding and Boyang Hong and Ming Zhang and Junzhe Wang and Senjie Jin and Enyu Zhou and Rui Zheng and Xiaoran Fan and Xiao Wang and Limao Xiong and Yuhao Zhou and Weiran Wang and Changhao Jiang and Yicheng Zou and Xiangyang Liu and Zhangyue Yin and Shihan Dou and Rongxiang Weng and Wensen Cheng and Qi Zhang and Wenjuan Qin and Yongyan Zheng and Xipeng Qiu and Xuanjing Huang and Tao Gui},
      year={2023},
      eprint={2309.07864},
      archivePrefix={arXiv},
      primaryClass={cs.AI},
      url={https://arxiv.org/abs/2309.07864}, 
}

@article{wang2024survey, title={A survey on large language model based autonomous agents}, volume={18}, ISSN={2095-2236}, url={http://dx.doi.org/10.1007/s11704-024-40231-1}, DOI={10.1007/s11704-024-40231-1}, number={6}, journal={Frontiers of Computer Science}, publisher={Springer Science and Business Media LLC}, author={Wang, Lei and Ma, Chen and Feng, Xueyang and Zhang, Zeyu and Yang, Hao and Zhang, Jingsen and Chen, Zhiyuan and Tang, Jiakai and Chen, Xu and Lin, Yankai and Zhao, Wayne Xin and Wei, Zhewei and Wen, Jirong}, year={2024}, month=Mar }

@inproceedings{shinn2023reflexion, series={NeurIPS 2023}, title={Reflexion: language agents with verbal reinforcement learning}, url={http://dx.doi.org/10.52202/075280-0377}, DOI={10.52202/075280-0377}, booktitle={Advances in Neural Information Processing Systems 36}, publisher={Neural Information Processing Systems Foundation, Inc. (NeurIPS)}, author={Shinn, Noah and Cassano, Federico and Gopinath, Ashwin and Narasimhan, Karthik and Yao, Shunyu}, year={2023}, pages={8634–8652}, collection={NeurIPS 2023} }

@inproceedings{kojima2022large, series={NeurIPS 2022}, title={Large Language Models Are Zero-Shot Reasoners}, url={http://dx.doi.org/10.52202/068431-1613}, DOI={10.52202/068431-1613}, booktitle={Advances in Neural Information Processing Systems 35}, publisher={Neural Information Processing Systems Foundation, Inc. (NeurIPS)}, author={Kojima, Takeshi and Gu, Shixiang (Shane) and Reid, Machel and Matsuo, Yutaka and Iwasawa, Yusuke}, year={2022}, pages={22199–22213}, collection={NeurIPS 2022} }

@inproceedings{lightman2024verify,
 author = {Lightman, Hunter and Kosaraju, Vineet and Burda, Yuri and Edwards, Harrison and Baker, Bowen and Lee, Teddy and Leike, Jan and Schulman, John  and Sutskever, Ilya and Cobbe, Karl},
 booktitle = {International Conference on Learning Representations},
 editor = {B. Kim and Y. Yue and S. Chaudhuri and K. Fragkiadaki and M. Khan and Y. Sun},
 pages = {39578--39601},
 title = {Let\textquotesingle s Verify Step by Step},
 url = {https://proceedings.iclr.cc/paper_files/paper/2024/file/aca97732e30bcf1303bc22ac3924fd16-Paper-Conference.pdf},
 volume = {2024},
 year = {2024}
}

@article{seung1992statistical, title={Statistical mechanics of learning from examples}, volume={45}, ISSN={1094-1622}, url={http://dx.doi.org/10.1103/PhysRevA.45.6056}, DOI={10.1103/physreva.45.6056}, number={8}, journal={Physical Review A}, publisher={American Physical Society (APS)}, author={Seung, H. S. and Sompolinsky, H. and Tishby, N.}, year={1992}, month=Apr, pages={6056–6091} }

@article{carleo2019machine, title={Machine learning and the physical sciences}, volume={91}, ISSN={1539-0756}, url={http://dx.doi.org/10.1103/RevModPhys.91.045002}, DOI={10.1103/revmodphys.91.045002}, number={4}, journal={Reviews of Modern Physics}, publisher={American Physical Society (APS)}, author={Carleo, Giuseppe and Cirac, Ignacio and Cranmer, Kyle and Daudet, Laurent and Schuld, Maria and Tishby, Naftali and Vogt-Maranto, Leslie and Zdeborová, Lenka}, year={2019}, month=Dec }

@inproceedings{christiano2017deep,
 author = {Christiano, Paul F and Leike, Jan and Brown, Tom and Martic, Miljan and Legg, Shane and Amodei, Dario},
 booktitle = {Advances in Neural Information Processing Systems},
 editor = {I. Guyon and U. Von Luxburg and S. Bengio and H. Wallach and R. Fergus and S. Vishwanathan and R. Garnett},
 pages = {},
 publisher = {Curran Associates, Inc.},
 title = {Deep Reinforcement Learning from Human Preferences},
 url = {https://proceedings.neurips.cc/paper_files/paper/2017/file/d5e2c0adad503c91f91df240d0cd4e49-Paper.pdf},
 volume = {30},
 year = {2017}
}

@misc{ziegler2019fine,
      title={Fine-Tuning Language Models from Human Preferences}, 
      author={Daniel M. Ziegler and Nisan Stiennon and Jeffrey Wu and Tom B. Brown and Alec Radford and Dario Amodei and Paul Christiano and Geoffrey Irving},
      year={2020},
      eprint={1909.08593},
      archivePrefix={arXiv},
      primaryClass={cs.CL},
      url={https://arxiv.org/abs/1909.08593}, 
}

@inproceedings{stiennon2020learning,
 author = {Stiennon, Nisan and Ouyang, Long and Wu, Jeffrey and Ziegler, Daniel and Lowe, Ryan and Voss, Chelsea and Radford, Alec and Amodei, Dario and Christiano, Paul F},
 booktitle = {Advances in Neural Information Processing Systems},
 editor = {H. Larochelle and M. Ranzato and R. Hadsell and M.F. Balcan and H. Lin},
 pages = {3008--3021},
 publisher = {Curran Associates, Inc.},
 title = {Learning to summarize with human feedback},
 url = {https://proceedings.neurips.cc/paper_files/paper/2020/file/1f89885d556929e98d3ef9b86448f951-Paper.pdf},
 volume = {33},
 year = {2020}
}

@inproceedings{ouyang2022training, series={NeurIPS 2022}, title={Training Language Models to Follow Instructions with Human Feedback}, url={http://dx.doi.org/10.52202/068431-2011}, DOI={10.52202/068431-2011}, booktitle={Advances in Neural Information Processing Systems 35}, publisher={Neural Information Processing Systems Foundation, Inc. (NeurIPS)}, author={Ouyang, Long and Wu, Jeffrey and Jiang, Xu and Almeida, Diogo and Wainwright, Carroll and Mishkin, Pamela and Zhang, Chong and Agarwal, Sandhini and Slama, Katarina and Ray, Alex and Schulman, John and Hilton, Jacob and Kelton, Fraser and Miller, Luke and Simens, Maddie and Askell, Amanda and Welinder, Peter and Christiano, Paul and Leike, Jan and Lowe, Ryan}, year={2022}, pages={27730–27744}, collection={NeurIPS 2022} }

@misc{bai2022training,
      title={Training a Helpful and Harmless Assistant with Reinforcement Learning from Human Feedback}, 
      author={Yuntao Bai and Andy Jones and Kamal Ndousse and Amanda Askell and Anna Chen and Nova DasSarma and Dawn Drain and Stanislav Fort and Deep Ganguli and Tom Henighan and Nicholas Joseph and Saurav Kadavath and Jackson Kernion and Tom Conerly and Sheer El-Showk and Nelson Elhage and Zac Hatfield-Dodds and Danny Hernandez and Tristan Hume and Scott Johnston and Shauna Kravec and Liane Lovitt and Neel Nanda and Catherine Olsson and Dario Amodei and Tom Brown and Jack Clark and Sam McCandlish and Chris Olah and Ben Mann and Jared Kaplan},
      year={2022},
      eprint={2204.05862},
      archivePrefix={arXiv},
      primaryClass={cs.CL},
      url={https://arxiv.org/abs/2204.05862}, 
}

@misc{bai2022constitutional,
      title={Constitutional {AI}: Harmlessness from {AI} Feedback}, 
      author={Yuntao Bai and Saurav Kadavath and Sandipan Kundu and Amanda Askell and Jackson Kernion and Andy Jones and Anna Chen and Anna Goldie and Azalia Mirhoseini and Cameron McKinnon and Carol Chen and Catherine Olsson and Christopher Olah and Danny Hernandez and Dawn Drain and Deep Ganguli and Dustin Li and Eli Tran-Johnson and Ethan Perez and Jamie Kerr and Jared Mueller and Jeffrey Ladish and Joshua Landau and Kamal Ndousse and Kamile Lukosuite and Liane Lovitt and Michael Sellitto and Nelson Elhage and Nicholas Schiefer and Noemi Mercado and Nova DasSarma and Robert Lasenby and Robin Larson and Sam Ringer and Scott Johnston and Shauna Kravec and Sheer El Showk and Stanislav Fort and Tamera Lanham and Timothy Telleen-Lawton and Tom Conerly and Tom Henighan and Tristan Hume and Samuel R. Bowman and Zac Hatfield-Dodds and Ben Mann and Dario Amodei and Nicholas Joseph and Sam McCandlish and Tom Brown and Jared Kaplan},
      year={2022},
      eprint={2212.08073},
      archivePrefix={arXiv},
      primaryClass={cs.CL},
      url={https://arxiv.org/abs/2212.08073}, 
}

@inproceedings{rafailov2023direct, series={NeurIPS 2023}, title={Direct Preference Optimization: Your Language Model is Secretly a Reward Model}, url={http://dx.doi.org/10.52202/075280-2338}, DOI={10.52202/075280-2338}, booktitle={Advances in Neural Information Processing Systems 36}, publisher={Neural Information Processing Systems Foundation, Inc. (NeurIPS)}, author={Rafailov, Rafael and Sharma, Archit and Mitchell, Eric and Manning, Christopher D and Ermon, Stefano and Finn, Chelsea}, year={2023}, pages={53728–53741}, collection={NeurIPS 2023} }

@misc{shao2024deepseekmath,
      title={{DeepSeekMath}: Pushing the Limits of Mathematical Reasoning in Open Language Models}, 
      author={Zhihong Shao and Peiyi Wang and Qihao Zhu and Runxin Xu and Junxiao Song and Xiao Bi and Haowei Zhang and Mingchuan Zhang and Y. K. Li and Y. Wu and Daya Guo},
      year={2024},
      eprint={2402.03300},
      archivePrefix={arXiv},
      primaryClass={cs.CL},
      url={https://arxiv.org/abs/2402.03300}, 
}

@article{guo2025deepseek, title={{DeepSeek}-{R1} incentivizes reasoning in {LLMs} through reinforcement learning}, volume={645}, ISSN={1476-4687}, url={http://dx.doi.org/10.1038/s41586-025-09422-z}, DOI={10.1038/s41586-025-09422-z}, number={8081}, journal={Nature}, publisher={Springer Science and Business Media LLC}, author={Guo, Daya and Yang, Dejian and Zhang, Haowei and Song, Junxiao and Wang, Peiyi and Zhu, Qihao and Xu, Runxin and Zhang, Ruoyu and Ma, Shirong and Bi, Xiao and Zhang, Xiaokang and Yu, Xingkai and Wu, Yu and Wu, Z. F. and Gou, Zhibin and Shao, Zhihong and Li, Zhuoshu and Gao, Ziyi and Liu, Aixin and Xue, Bing and Wang, Bingxuan and Wu, Bochao and Feng, Bei and Lu, Chengda and Zhao, Chenggang and Deng, Chengqi and Ruan, Chong and Dai, Damai and Chen, Deli and Ji, Dongjie and Li, Erhang and Lin, Fangyun and Dai, Fucong and Luo, Fuli and Hao, Guangbo and Chen, Guanting and Li, Guowei and Zhang, H. and Xu, Hanwei and Ding, Honghui and Gao, Huazuo and Qu, Hui and Li, Hui and Guo, Jianzhong and Li, Jiashi and Chen, Jingchang and Yuan, Jingyang and Tu, Jinhao and Qiu, Junjie and Li, Junlong and Cai, J. L. and Ni, Jiaqi and Liang, Jian and Chen, Jin and Dong, Kai and Hu, Kai and You, Kaichao and Gao, Kaige and Guan, Kang and Huang, Kexin and Yu, Kuai and Wang, Lean and Zhang, Lecong and Zhao, Liang and Wang, Litong and Zhang, Liyue and Xu, Lei and Xia, Leyi and Zhang, Mingchuan and Zhang, Minghua and Tang, Minghui and Zhou, Mingxu and Li, Meng and Wang, Miaojun and Li, Mingming and Tian, Ning and Huang, Panpan and Zhang, Peng and Wang, Qiancheng and Chen, Qinyu and Du, Qiushi and Ge, Ruiqi and Zhang, Ruisong and Pan, Ruizhe and Wang, Runji and Chen, R. J. and Jin, R. L. and Chen, Ruyi and Lu, Shanghao and Zhou, Shangyan and Chen, Shanhuang and Ye, Shengfeng and Wang, Shiyu and Yu, Shuiping and Zhou, Shunfeng and Pan, Shuting and Li, S. S. and Zhou, Shuang and Wu, Shaoqing and Yun, Tao and Pei, Tian and Sun, Tianyu and Wang, T. and Zeng, Wangding and Liu, Wen and Liang, Wenfeng and Gao, Wenjun and Yu, Wenqin and Zhang, Wentao and Xiao, W. L. and An, Wei and Liu, Xiaodong and Wang, Xiaohan and Chen, Xiaokang and Nie, Xiaotao and Cheng, Xin and Liu, Xin and Xie, Xin and Liu, Xingchao and Yang, Xinyu and Li, Xinyuan and Su, Xuecheng and Lin, Xuheng and Li, X. Q. and Jin, Xiangyue and Shen, Xiaojin and Chen, Xiaosha and Sun, Xiaowen and Wang, Xiaoxiang and Song, Xinnan and Zhou, Xinyi and Wang, Xianzu and Shan, Xinxia and Li, Y. K. and Wang, Y. Q. and Wei, Y. X. and Zhang, Yang and Xu, Yanhong and Li, Yao and Zhao, Yao and Sun, Yaofeng and Wang, Yaohui and Yu, Yi and Zhang, Yichao and Shi, Yifan and Xiong, Yiliang and He, Ying and Piao, Yishi and Wang, Yisong and Tan, Yixuan and Ma, Yiyang and Liu, Yiyuan and Guo, Yongqiang and Ou, Yuan and Wang, Yuduan and Gong, Yue and Zou, Yuheng and He, Yujia and Xiong, Yunfan and Luo, Yuxiang and You, Yuxiang and Liu, Yuxuan and Zhou, Yuyang and Zhu, Y. X. and Huang, Yanping and Li, Yaohui and Zheng, Yi and Zhu, Yuchen and Ma, Yunxian and Tang, Ying and Zha, Yukun and Yan, Yuting and Ren, Z. Z. and Ren, Zehui and Sha, Zhangli and Fu, Zhe and Xu, Zhean and Xie, Zhenda and Zhang, Zhengyan and Hao, Zhewen and Ma, Zhicheng and Yan, Zhigang and Wu, Zhiyu and Gu, Zihui and Zhu, Zijia and Liu, Zijun and Li, Zilin and Xie, Ziwei and Song, Ziyang and Pan, Zizheng and Huang, Zhen and Xu, Zhipeng and Zhang, Zhongyu and Zhang, Zhen}, year={2025}, month=Sep, pages={633–638} }

@InProceedings{haarnoja2018soft,
  title = 	 {Soft Actor-Critic: Off-Policy Maximum Entropy Deep Reinforcement Learning with a Stochastic Actor},
  author =       {Haarnoja, Tuomas and Zhou, Aurick and Abbeel, Pieter and Levine, Sergey},
  booktitle = 	 {Proceedings of the 35th International Conference on Machine Learning},
  pages = 	 {1861--1870},
  year = 	 {2018},
  editor = 	 {Dy, Jennifer and Krause, Andreas},
  volume = 	 {80},
  series = 	 {Proceedings of Machine Learning Research},
  month = 	 {10--15 Jul},
  publisher =    {PMLR},
  url = 	 {https://proceedings.mlr.press/v80/haarnoja18b.html}
}

@article{jaynes1957information, title={Information Theory and Statistical Mechanics}, volume={106}, ISSN={0031-899X}, url={http://dx.doi.org/10.1103/PhysRev.106.620}, DOI={10.1103/physrev.106.620}, number={4}, journal={Physical Review}, publisher={American Physical Society (APS)}, author={Jaynes, E. T.}, year={1957}, month=May, pages={620–630} }

@inproceedings{schaeffer2023emergent, series={NeurIPS 2023}, title={Are Emergent Abilities of Large Language Models a Mirage?}, url={http://dx.doi.org/10.52202/075280-2425}, DOI={10.52202/075280-2425}, booktitle={Advances in Neural Information Processing Systems 36}, publisher={Neural Information Processing Systems Foundation, Inc. (NeurIPS)}, author={Schaeffer, Rylan and Miranda, Brando and Koyejo, Sanmi}, year={2023}, pages={55565–55581}, collection={NeurIPS 2023} }

@article{schnakenberg1976network, title={Network theory of microscopic and macroscopic behavior of master equation systems}, volume={48}, ISSN={0034-6861}, url={http://dx.doi.org/10.1103/RevModPhys.48.571}, DOI={10.1103/revmodphys.48.571}, number={4}, journal={Reviews of Modern Physics}, publisher={American Physical Society (APS)}, author={Schnakenberg, J.}, year={1976}, month=Oct, pages={571–585} }

@book{sutton2018reinforcement,
  title     = {Reinforcement Learning: An Introduction},
  author    = {Sutton, Richard S. and Barto, Andrew G.},
  year      = {2018},
  publisher = {MIT Press},
  address   = {Cambridge, MA},
  note      = {A Bradford Book},
  url       = {https://mitpress.mit.edu/9780262039246/reinforcement-learning/},
}

\endgroup
\clearpage
\onecolumn
\appendix
\setcounter{table}{0}
\renewcommand{\thetable}{\thesection\arabic{table}}
\section{Complete prompts and candidate mappings}\label{app:prompts}

This appendix provides the complete inputs used in the T1 and T2 main experiments. For each task, we first show the complete prompt used for one source state, followed by all state descriptions and source-specific candidate mappings. To reconstruct inputs for the other source states, only the current-state description and the destination text following each candidate code need to be replaced; all other characters, message roles and their order remain unchanged. State identifiers in the tables serve only as indices.

\subsection{Chat templates and inference modes}

T1 contains only a user message, whereas T2 also contains a system message. The complete inputs below all use direct-answer mode. Task messages are identical in the two modes; the only difference is the assistant prefix appended after the messages. Direct-answer mode inserts a closed, empty thinking block, whereas thinking mode lets the model continue an unclosed thinking block.
\begingroup
\tcbset{equal height group=assistant-prefix-pair,valign=top}
\par\noindent\begin{minipage}[t]{0.48\linewidth}\vspace{0pt}
\begin{promptbox}{Direct answer: assistant prefix}
\begin{PromptText}
<|im_start|>assistant
<think>

</think>

\end{PromptText}
\end{promptbox}
\end{minipage}\hfill\begin{minipage}[t]{0.48\linewidth}\vspace{0pt}
\begin{promptbox}{Thinking: assistant prefix}
\begin{PromptText}
<|im_start|>assistant
<think>
\end{PromptText}
\end{promptbox}
\end{minipage}\par\medskip
\endgroup
The generation limit is 32 tokens in direct-answer mode and 16,384 tokens in thinking mode, including reasoning and the final answer. In the main experiments, complete T1 inputs are all 296 tokens long, whereas T2 inputs contain 253--266 tokens; the thinking template reduces each input by 2 tokens. These lengths include chat-template markers, and all inputs are retained in full.

\clearpage
\subsection{T1: safety-test sequencing}\label{app:prompt-t1}

The representative state in the main text is \(t5\); the box below shows its complete input. The meanings of candidate codes A--G vary across source states; the full mappings are given in Table~\ref{tab:prompt-map-t1}.
\begin{promptbox}{Complete T1 input (source state \texttt{t5})}
\begin{PromptText}
<|im_start|>user
A security team uses synthetic data to test whether an AI office assistant follows malicious instructions embedded in untrusted content. A test case has three attributes: delivery carrier (inbound email or retrieved document), persuasion pretext (time pressure or claimed authority), and probed behavior (create a local draft or attempt an external send). All tests use inert placeholders, a mock tool and no real recipients or private data. These are test-case categories, not instructions to carry out an attack. The generator proposes test cases; a separate harness checks whether the assistant respects the trust boundary.
The current test is: carrier=retrieved document; pretext=time pressure; probe=external send
Possible next test cases:
A: carrier=retrieved document; pretext=claimed authority; probe=local draft
B: carrier=inbound email; pretext=claimed authority; probe=local draft
C: carrier=inbound email; pretext=time pressure; probe=external send
D: carrier=inbound email; pretext=time pressure; probe=local draft
E: carrier=retrieved document; pretext=claimed authority; probe=external send
F: carrier=retrieved document; pretext=time pressure; probe=local draft
G: carrier=inbound email; pretext=claimed authority; probe=external send
Propose one plausible next test case from the candidates. Do not repeat the current test. Reply only ANSWER: <ID>.<|im_end|>
<|im_start|>assistant
<think>

</think>

\end{PromptText}
\end{promptbox}

\clearpage
\begingroup\small
\renewcommand{\arraystretch}{1.15}
\setlength{\LTcapwidth}{\linewidth}
\begin{longtable}{@{}>{\ttfamily}p{0.07\linewidth}>{\raggedright\arraybackslash\ttfamily}p{\dimexpr0.93\linewidth-2\tabcolsep\relax}@{}}
\caption{T1 state descriptions.}\label{tab:prompt-states-t1}\\
\toprule
\normalfont State & \normalfont State description (original text) \\
\midrule
\endfirsthead
\toprule
\normalfont State & \normalfont State description (continued) \\
\midrule
\endhead
t0 & carrier=inbound email; pretext=time pressure; probe=local draft \\
t1 & carrier=inbound email; pretext=time pressure; probe=external send \\
t2 & carrier=inbound email; pretext=claimed authority; probe=local draft \\
t3 & carrier=inbound email; pretext=claimed authority; probe=external send \\
t4 & carrier=retrieved document; pretext=time pressure; probe=local draft \\
t5 & carrier=retrieved document; pretext=time pressure; probe=external send \\
t6 & carrier=retrieved document; pretext=claimed authority; probe=local draft \\
t7 & carrier=retrieved document; pretext=claimed authority; probe=external send \\
\bottomrule
\end{longtable}
\endgroup
\begingroup\small
\renewcommand{\arraystretch}{1.12}
\setlength{\LTcapwidth}{\linewidth}
\begin{longtable}{@{}>{\raggedright\arraybackslash}p{0.1\linewidth}*{7}{>{\centering\arraybackslash}p{\dimexpr(0.9\linewidth-14\tabcolsep)/7\relax}}@{}}
\caption{Source-specific candidate-code mappings for T1. Each cell gives the destination state \(j\) (utility \(u_j\)).}\label{tab:prompt-map-t1}\\
\toprule
Source state & \texttt{A} & \texttt{B} & \texttt{C} & \texttt{D} & \texttt{E} & \texttt{F} & \texttt{G} \\
\midrule
\endfirsthead
\toprule
Source state & \texttt{A} & \texttt{B} & \texttt{C} & \texttt{D} & \texttt{E} & \texttt{F} & \texttt{G} \\
\midrule
\endhead
\texttt{t0} & \texttt{t2} (\(0.25\)) & \texttt{t1} (\(0.5\)) & \texttt{t3} (\(0.75\)) & \texttt{t5} (\(0.75\)) & \texttt{t6} (\(0.5\)) & \texttt{t7} (\(1\)) & \texttt{t4} (\(0.25\)) \\
\texttt{t1} & \texttt{t7} (\(1\)) & \texttt{t5} (\(0.75\)) & \texttt{t0} (\(0\)) & \texttt{t6} (\(0.5\)) & \texttt{t3} (\(0.75\)) & \texttt{t2} (\(0.25\)) & \texttt{t4} (\(0.25\)) \\
\texttt{t2} & \texttt{t7} (\(1\)) & \texttt{t3} (\(0.75\)) & \texttt{t0} (\(0\)) & \texttt{t1} (\(0.5\)) & \texttt{t6} (\(0.5\)) & \texttt{t4} (\(0.25\)) & \texttt{t5} (\(0.75\)) \\
\texttt{t3} & \texttt{t4} (\(0.25\)) & \texttt{t7} (\(1\)) & \texttt{t2} (\(0.25\)) & \texttt{t0} (\(0\)) & \texttt{t1} (\(0.5\)) & \texttt{t6} (\(0.5\)) & \texttt{t5} (\(0.75\)) \\
\texttt{t4} & \texttt{t2} (\(0.25\)) & \texttt{t3} (\(0.75\)) & \texttt{t1} (\(0.5\)) & \texttt{t7} (\(1\)) & \texttt{t0} (\(0\)) & \texttt{t5} (\(0.75\)) & \texttt{t6} (\(0.5\)) \\
\texttt{t5} & \texttt{t6} (\(0.5\)) & \texttt{t2} (\(0.25\)) & \texttt{t1} (\(0.5\)) & \texttt{t0} (\(0\)) & \texttt{t7} (\(1\)) & \texttt{t4} (\(0.25\)) & \texttt{t3} (\(0.75\)) \\
\texttt{t6} & \texttt{t7} (\(1\)) & \texttt{t3} (\(0.75\)) & \texttt{t2} (\(0.25\)) & \texttt{t4} (\(0.25\)) & \texttt{t0} (\(0\)) & \texttt{t5} (\(0.75\)) & \texttt{t1} (\(0.5\)) \\
\texttt{t7} & \texttt{t4} (\(0.25\)) & \texttt{t1} (\(0.5\)) & \texttt{t6} (\(0.5\)) & \texttt{t5} (\(0.75\)) & \texttt{t3} (\(0.75\)) & \texttt{t0} (\(0\)) & \texttt{t2} (\(0.25\)) \\
\bottomrule
\end{longtable}
\endgroup

\paragraph{Edge-specific reward residual.}
The reward ablation retains the T1 states, candidate-code mappings and \(u_j\), and assigns reward \(1+u_j+\alpha\epsilon_{ij}\) to valid transitions. For state identifiers \(t_i,t_j\), define \(d_{ij}=(i-j)\bmod 8\). For \(i\ne j\),
\begin{equation}\label{eq:app-edge-epsilon}
\epsilon_{ij}=\begin{cases}
 1, & d_{ij}\in\{1,2,3\},\\
 0, & d_{ij}=4,\\
-1, & d_{ij}\in\{5,6,7\}.
\end{cases}
\end{equation}
Self-transitions are excluded; \(\epsilon_{ij}\) sums to zero in every row and column.

\clearpage
\subsection{T2: sequencing physics-auditing steps}\label{app:prompt-t2}

The box below shows the complete input for source state \(p0\), including the system and user messages. Codes A--D correspond to the two neighbours in either direction from this source on the ring.
\begin{promptbox}{Complete T2 input (source state \texttt{p0})}
\begin{PromptText}
<|im_start|>system
You are a physics-audit mutator. Output exactly ANSWER: followed by one uppercase ID A-D. Never explain.<|im_end|>
<|im_start|>user
You are evaluating a synthetic AI-physics reasoning benchmark. A planar two-layer wall has layer 1 thickness L1 and conductivity k1, layer 2 thickness L2 and conductivity k2, and a finite thermal contact resistance Rc at the interface. The outer faces are held at Th and Tc with Th>Tc; there is steady one-dimensional conduction, no internal heat generation, and no convection boundary unless explicitly stated. The generator proposes the next short audit checkpoint; a separate rubric checks physical consistency. Do not carry out an experiment or use real data.

Current checkpoint: state the geometry, sign convention and no-source steady assumptions
Possible next checkpoints:
A: check the limiting case Rc approaching zero
B: flag a tempting error: forcing equal interface temperature or adding an unstated convection term
C: draw the series thermal-resistance path including Rc
D: write the energy balance showing the heat flux is constant
Propose the next checkpoint that most usefully audits the current step. Do not repeat the current checkpoint. Reply only ANSWER: <ID>.<|im_end|>
<|im_start|>assistant
<think>

</think>

\end{PromptText}
\end{promptbox}

\clearpage
\begingroup\small
\renewcommand{\arraystretch}{1.15}
\setlength{\LTcapwidth}{\linewidth}
\begin{longtable}{@{}>{\ttfamily}p{0.07\linewidth}>{\raggedright\arraybackslash\ttfamily}p{\dimexpr0.93\linewidth-2\tabcolsep\relax}@{}}
\caption{T2 state descriptions.}\label{tab:prompt-states-t2}\\
\toprule
\normalfont State & \normalfont State description (original text) \\
\midrule
\endfirsthead
\toprule
\normalfont State & \normalfont State description (continued) \\
\midrule
\endhead
p0 & state the geometry, sign convention and no-source steady assumptions \\
p1 & draw the series thermal-resistance path including Rc \\
p2 & write the energy balance showing the heat flux is constant \\
p3 & write Fourier's relation across layer 1 \\
p4 & write Fourier's relation across layer 2 \\
p5 & write the temperature jump relation at the contact resistance \\
p6 & apply the hot-face and cold-face boundary conditions \\
p7 & solve the signed heat flux in terms of both layer resistances and Rc \\
p8 & solve the interface temperature from the same flux \\
p9 & check units and dimensions of the total resistance and heat flux \\
p10 & check the limiting case Rc approaching zero \\
p11 & flag a tempting error: forcing equal interface temperature or adding an unstated convection term \\
\bottomrule
\end{longtable}
\endgroup
\begingroup\small
\renewcommand{\arraystretch}{1.12}
\setlength{\LTcapwidth}{\linewidth}
\begin{longtable}{@{}>{\raggedright\arraybackslash}p{0.1\linewidth}*{4}{>{\centering\arraybackslash}p{\dimexpr(0.9\linewidth-8\tabcolsep)/4\relax}}@{}}
\caption{Source-specific candidate-code mappings for T2. Each cell gives the destination state \(j\) (utility \(u_j\)).}\label{tab:prompt-map-t2}\\
\toprule
Source state & \texttt{A} & \texttt{B} & \texttt{C} & \texttt{D} \\
\midrule
\endfirsthead
\toprule
Source state & \texttt{A} & \texttt{B} & \texttt{C} & \texttt{D} \\
\midrule
\endhead
\texttt{p0} & \texttt{p10} (\(0.9\)) & \texttt{p11} (\(0.1\)) & \texttt{p1} (\(0.65\)) & \texttt{p2} (\(0.75\)) \\
\texttt{p1} & \texttt{p3} (\(0.45\)) & \texttt{p11} (\(0.1\)) & \texttt{p0} (\(0.35\)) & \texttt{p2} (\(0.75\)) \\
\texttt{p2} & \texttt{p3} (\(0.45\)) & \texttt{p4} (\(0.45\)) & \texttt{p0} (\(0.35\)) & \texttt{p1} (\(0.65\)) \\
\texttt{p3} & \texttt{p2} (\(0.75\)) & \texttt{p1} (\(0.65\)) & \texttt{p4} (\(0.45\)) & \texttt{p5} (\(0.9\)) \\
\texttt{p4} & \texttt{p6} (\(0.7\)) & \texttt{p2} (\(0.75\)) & \texttt{p3} (\(0.45\)) & \texttt{p5} (\(0.9\)) \\
\texttt{p5} & \texttt{p7} (\(1\)) & \texttt{p6} (\(0.7\)) & \texttt{p4} (\(0.45\)) & \texttt{p3} (\(0.45\)) \\
\texttt{p6} & \texttt{p4} (\(0.45\)) & \texttt{p5} (\(0.9\)) & \texttt{p7} (\(1\)) & \texttt{p8} (\(0.85\)) \\
\texttt{p7} & \texttt{p8} (\(0.85\)) & \texttt{p5} (\(0.9\)) & \texttt{p6} (\(0.7\)) & \texttt{p9} (\(0.95\)) \\
\texttt{p8} & \texttt{p9} (\(0.95\)) & \texttt{p10} (\(0.9\)) & \texttt{p6} (\(0.7\)) & \texttt{p7} (\(1\)) \\
\texttt{p9} & \texttt{p7} (\(1\)) & \texttt{p11} (\(0.1\)) & \texttt{p8} (\(0.85\)) & \texttt{p10} (\(0.9\)) \\
\texttt{p10} & \texttt{p11} (\(0.1\)) & \texttt{p8} (\(0.85\)) & \texttt{p9} (\(0.95\)) & \texttt{p0} (\(0.35\)) \\
\texttt{p11} & \texttt{p10} (\(0.9\)) & \texttt{p9} (\(0.95\)) & \texttt{p0} (\(0.35\)) & \texttt{p1} (\(0.65\)) \\
\bottomrule
\end{longtable}
\endgroup

\end{document}